\pdfoutput=1

\documentclass[11pt]{article}

\usepackage[final]{acl}

\usepackage{times}
\usepackage{latexsym}

\usepackage{titletoc}

\usepackage{amsmath,amsfonts,bm}

\def\eqref#1{equation~\ref{#1}}

\def\1{\bm{1}}

\DeclareMathAlphabet{\mathsfit}{\encodingdefault}{\sfdefault}{m}{sl}
\SetMathAlphabet{\mathsfit}{bold}{\encodingdefault}{\sfdefault}{bx}{n}

\usepackage{hyperref}
\usepackage{url}
\usepackage{comment}
\usepackage{graphicx}
\usepackage{pifont}
\usepackage{enumitem}
\usepackage{amssymb}

\usepackage[T1]{fontenc}

\usepackage[utf8]{inputenc}
\usepackage{tabularray}

\usepackage{microtype}

\usepackage{inconsolata}

\usepackage{graphicx}

\usepackage{array}
\usepackage{bbm}
\usepackage{graphicx}
\usepackage{makecell}
\usepackage{xcolor}
\usepackage{soul}
\usepackage{xspace}
\usepackage{multirow}
\usepackage{booktabs}
\newcolumntype{P}[1]{>{\centering\arraybackslash}p{#1}}

\usepackage[most]{tcolorbox}
\AtBeginEnvironment{tcolorbox}{\small}

\title{Instructions for *ACL Proceedings}

\author{First Author \\
  Affiliation / Address line 1 \\
  Affiliation / Address line 2 \\
  Affiliation / Address line 3 \\
  \texttt{email@domain} \\\And
  Second Author \\
  Affiliation / Address line 1 \\
  Affiliation / Address line 2 \\
  Affiliation / Address line 3 \\
  \texttt{email@domain} \\}

\usepackage{array}
\newcommand{\PreserveBackslash}[1]{\let\temp=\\#1\let\\=\temp}
\newcolumntype{C}[1]{>{\PreserveBackslash\centering}p{#1}}
\newcolumntype{R}[1]{>{\PreserveBackslash\raggedleft}p{#1}}
\newcolumntype{L}[1]{>{\PreserveBackslash\raggedright}p{#1}}

\title{ICL Robustnes to Vocab Shuffling}
\title{ICL can reverse-engineer (or, recover) bijective word shufflings }
\title{LLMs Can In-context Recover Novel Bijections Word Shufflings}
\title{LLMs Can In-context Learn novel Bijective Word Shuffling}
\title{LLM Caesar Cipher: In-Context Learning Recovers Bijective Word Shuffling}
\title{Quantifying ``Learning'' in In-Context Learning via Language Cipher Problems}
\title{Quantifying ``Learning'' in In-Context Learning: \\  A Case Study on Language Cipher Problem}
\title{How much ``Learning'' Happens In-Context? \\  A Case Study on Language Cipher Problem}
\title{Measuring ``Learning'' in In-Context Learning via Language Cipher Problems}
\title{How much ``Learning'' is in In-Context Learning? \\  A Case Study on Language Cipher Problem}
\title{How much does In-Context Learning ``Learn''? \\  A Case Study on Language Cipher Problem}
\title{Quantifying ``Learning'' in In-Context Learning \\  via Language Ciphers}
\title{\name: Quantifying ``Learning'' in In-Context Learning \\ via Substitution Ciphers}

\author{Zhouxiang Fang,  \; Aayush Mishra,  \; Muhan Gao, \; Anqi Liu \textnormal{ and } Daniel Khashabi\\
Department of Computer Science\\
Johns Hopkins University\\
Baltimore, MD 21218, USA 
}

\def\mystrut(#1,#2){\vrule height #1pt depth #2pt width 0pt}   

\definecolor{purple}{rgb}{0.5,0,1}
\definecolor{dcyan}{rgb}{0.2,0.6,0.5}
\definecolor{DarkGreen}{RGB}{51,140,0}
\definecolor{light-gray}{gray}{0.95} 

\definecolor{darkgreen}{RGB}{0,140,0}
\definecolor{darkred}{RGB}{200,0,0}
\definecolor{DarkRed}{RGB}{200,0,0}
\definecolor{lightgreen}{RGB}{189,252,192}
\definecolor{lightred}{RGB}{255,205,212}
\definecolor{lightyellow}{RGB}{255,240,160}
\definecolor{lightblue}{RGB}{195,221,255}
\definecolor{lightpurple}{RGB}{232,209,255}

\newcommand{\name}{\textsc{ICL Ciphers}}
\newcommand{\gemma}{Gemma 2 }
\newcommand{\llama}{Llama 3.1 }
\newcommand{\qwen}{QWen 2.5 }
\newcommand{\olmo}{OLMo }
\begin{document}

\maketitle

\begin{abstract}
Recent works have suggested that In-Context Learning (ICL) operates in dual modes, i.e. task retrieval (remember learned patterns from pre-training) and task learning (inference-time ``learning" from demonstrations). However, disentangling these the two modes remains a challenging goal. 
We introduce \name, a class of task reformulations based on \emph{substitution ciphers} borrowed from classic cryptography.
In this approach, a subset of tokens in the in-context inputs are substituted with other (irrelevant)
tokens, rendering English sentences less comprehensible to human eye. 
However, by design, \emph{there is a latent, fixed pattern to this substitution, making it reversible}. 
This bijective (reversible) cipher ensures that
the task remains a well-defined task in some abstract sense, despite the transformations. 
It is a curious question if LLMs can solve tasks reformulated by \name{} with a \textsc{Bijective} mapping, which requires ``deciphering'' the latent cipher. 
We show that LLMs are better at solving tasks reformulated by \name{} with \textsc{Bijective} mappings than the \textsc{Non-Bijective} (irreversible) baseline,  
providing a novel approach to quantify ``learning'' in ICL. 
While this gap is small, it is consistent across the board on four datasets and six models. 
Finally, our interpretability analysis shows evidence that LLMs can internally decode ciphered inputs.\footnote{Our code is available at this 
\href{https://github.com/jhu-CLSP/icl-ciphers}{repository}.
} 
\end{abstract}

\section{Introduction}
\label{sec:intro}

\begin{figure}[t]
\includegraphics[width=17.3cm,trim=3.2cm 2.8cm 3cm 0.1cm,clip=true]{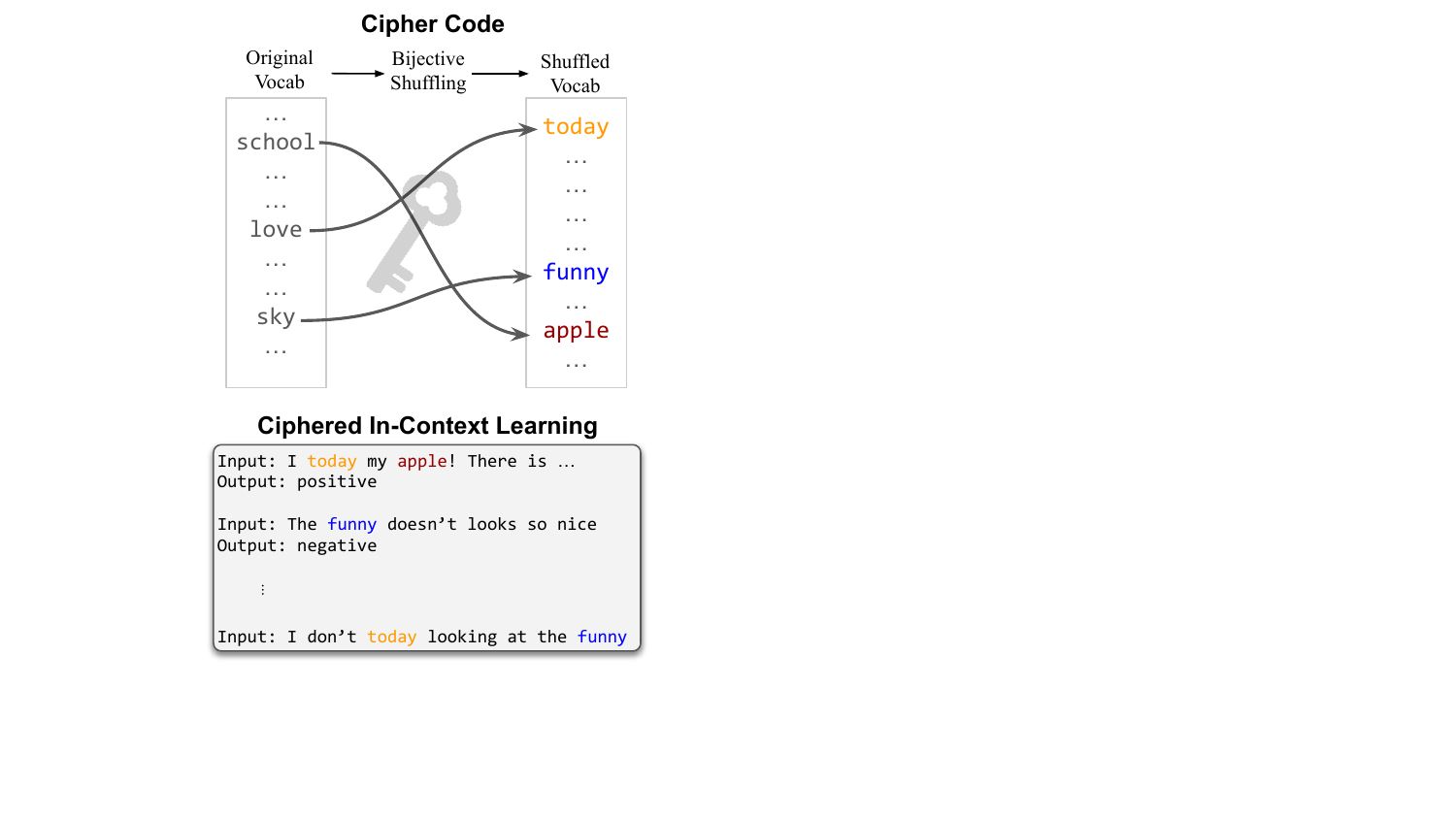}
\caption{
An example of \name, a cryptographic task reformulation where a subset of tokens are ciphered (replaced with other tokens in the lexicon) via a \textsc{Bijective} mapping (e.g., each instance of ``school''  is replaced with ``apple''.) 
Since this cipher is a bijection, one can recover the original format of the ICL instance, ensuring the well-defined task upon the transformations. 
}
\label{fig:pipeline}
\end{figure}
In-Context Learning (ICL) is an emergent behavior in Large Language Models (LLMs) that allows them to identify patterns in demonstrations given as prompts and apply these patterns to similar tasks~\citep{brown2020language}. 
This intriguing inference-time ability has prompted many studies.
Despite recent efforts~\cite[inter alia]{min2022rethinking,srivastava2023beyond,shen2024icl_vs_gd}, the literature's understanding of the functional aspects of ICL remains elusive and contentious. 


Most pertinent to our study, \citet{pan2023incontext,lin2024dual,wang2024investigating} propose ICL's dual behavior: \textbf{\emph{task retrieval}} (TR), which involves recalling a previously encountered task from pre-training data through its demonstrations, and \textbf{\emph{task learning}} (TL), which refers to the ability to grasp new input-label mappings that were \emph{not} seen during pre-training. 
Although these two mechanisms are not necessarily separate in practice, examining them independently may help researchers better understand their strengths and limitations. 
This distinction is important as TL reflects whether models can generalize to truly new tasks or label spaces from just a few examples, which is the assumption of many practical uses of ICL.
However, since most of the existing tasks are already included in pretraining, it is non-trivial to find new tasks during inference and measure TL independently. 
\citet{pan2023incontext} measure TL by assessing task performance when labels are substituted with abstract symbols (such as numbers or letters) that have never co-occurred with the inputs during pre-training. 
However, TR may partially influence this strategy.
LLMs could still use the intact human-readable inputs and prompt structure to deduce the task, thereby performing implicit task retrieval. 
This consideration motivates the exploration of alternative approaches for quantifying task learning.

In this study, we introduce \textbf{\name}, 
a class of prompt reformulations based on \emph{substitution ciphers} borrowed from cryptography, applied to task \textit{inputs}.
For example, in a sentiment classification task, we apply \textsc{Bijective} shuffling to part of the LLM's original vocabulary, ensuring a one-to-one correspondence between tokens in the shuffled and original vocabularies. 
We then replace tokens in the input text with their corresponding tokens based on this mapping (e.g., every instance of ``love'' is replaced with ``today''; see~Fig.\ref{fig:pipeline}).

The outcome of substitution ciphers is generally not easily interpretable by humans (see Fig.\ref{fig:pipeline} for examples), resembling a random shuffling of words. However, since ICL ciphers are \emph{reversible}, the original tasks can be reconstructed from the encoded version, ensuring that the task, 
still remains learnable. 
This lack of interpretability is a design feature (rather than a flaw) here as it greatly reduces the likelihood that our prompts have been encountered in the pre-training data.
As a result, our working hypothesis is that any gains above the \textsc{Non-Bijective} shuffles should be indicative of TL (as opposed to TR) within ICL. Unlike previous works \cite{pan2023incontext,wang2024investigating} that intervene in task outputs through label shuffling, our approach modifies task inputs. This creates instances less likely to have been encountered in pre-training data.

In summary, we evaluate \name~ using six models of different sizes across four well-known benchmarks and different few-shot numbers. 
Our empirical results demonstrate that ICL achieves better-than-random performance on ciphered tasks (\S\ref{results:llm:solve:it}). 
For example, on the \textsc{Bijective} ciphered Amazon dataset, Llama3.1 (8B) averages 7.1\% higher accuracy than \textsc{Non-Bijective} ciphers, across various demonstration counts (\autoref{tab:llama8b_infor_sample}). 
This suggests that LLMs can learn and decode these random bijections, enabling them to solve ICL Ciphers. 
Furthermore, we provide additional results with the shuffling rate and model scale. 
Finally, we perform an interpretability analysis (\S\ref{sec:probing})  which reveals promising, albeit weak, trends in their ability to decode the ciphered inputs.



\section{Defining \name{}}
\label{sec:framework}

\subsection{Preliminaries: In-Context Learning}
Let $f_\theta$ denote a pre-trained language model parameterized by $\theta$. This model performs ICL by conditioning on an ordered set of $n$-many input-output pairs $D_\text{demo}= (x_1, y_1, x_2, y_2, \hdots, x_n, y_n)$.
To measure this model's competence, we evaluate it on a collection of input-output pairs $D_\text{test} = \{(x_i, y_i)\}$. Specifically, for instance $(x_\text{test}, y_\text{test}) \sim D_\text{test}$, from an LM conditioned on the demonstrations with an appropriate encoding:
$y_\text{pred}\sim f_\theta(D_\text{demo}, x_\text{test})$ we extract a predicted label $y_\text{pred}$ which is then compared against the gold label  $y_\text{test}$.

\subsection{\name{}}
\label{subsec:iclciphers}
A simple substitution cipher is a technique for encoding messages.
Specifically, each letter in the plain text is substituted with a
different letter from the alphabet, usually according to a predetermined
mapping or key.
\name{} are token-level substitution ciphers that are applied to
demonstration inputs in ICL.
Formally, we define a ICL cipher \( c: V \rightarrow V \) that maps
each token in the lexicon \( V = \{t_j\}_{j=1}^{|V|} \) to another token.
Note that a token is allowed to be mapped to itself.
If all the tokens are mapped to themselves (i.e., \( c(t_j) = t_j \) for all \( j \)), then the ICL cipher is equal to an identity function, and substitution with this mapping would lead to no changes in the text.
We define the tokens that are mapped to \emph{different} tokens as ciphered tokens $S := \{ t_j | t_j \in V, c(t_j) \neq t_j \}$.
 The proportion of shuffled tokens in the lexicon is called \textit{shuffle rate} $r \in [0, 1]$.
The mapping of ciphered tokens depends on the specific type of \name{}, which we discuss next.

\subsection{\textsc{Bijective} ciphers}
\label{subsec:bijectiveciphers}
We create a \textsc{Bijective} mapping between two permuted orders of $S$.
For example, say the token ``school'' is mapped to ``apple'', as illustrated in~Fig.\ref{fig:pipeline}.
Let the input $x_i$ be constituted of $K_i$ tokens, i.e., $x_i$ is the ordered sequence of tokens $(t_1, \hdots, t_{K_i})$.
For all $t_j = \text{school} \in x_i$ or $x_\text{test}, c(t_j) = \text{apple}$. This results in corresponding ciphered inputs $x_i'$ or $x_\text{test}'$. Moreover, as $c$ is a bijection, $\exists\hspace{2pt}c^{-1}$ such that for all $t_j = \text{apple} \in x_i'$ or $x_\text{test}', c^{-1}(t_j) = \text{school}$.
Note that ``apple'' doesn't have to be mapped back to ``school''.

\paragraph{Decipherability of \textsc{Bijective} cipher:}
Since we ensure the mapping is \textsc{Bijective} (reversible), theoretically the models can learn the mapping through enough demonstrations.
Let the actual function between all $(x_i, y_i)$ pairs be $h$, i.e. $h(x_i) = y_i, \forall (x_i, y_i) \in D_\text{demo} \cup D_\text{test}$. Using ICL, the model $f_\theta$ employs both TR and TL to approximate $h' \approx h$ such that $h'(x_i) \approx y_i$. This original function $h$ cannot be expected to work on ciphered (or shuffled) inputs $x_i'$. However, there is a corresponding function $g(x_i') = h(c^{-1}(x_i'))$ that is equivalent to $h(x_i)$.
This shows that $h$ is still recoverable from the ciphered inputs. In natural language, replacing a word with another fixed but randomly decided word can completely change the meaning of its context. Any TR capabilities are expected to be severely hurt with ciphered inputs. To perform well on $D_\text{test}$, the model must rely heavily on TL to learn and perform this composite function.

\subsection{\textsc{Non-Bijective} Ciphers}
\label{subsec:non-bijectiveciphers}

For comparison with \textsc{Bijective} ciphers~(\S\ref{subsec:bijectiveciphers}), we also create a \textsc{Non-Bijective} cipher.
In this cipher, whenever a token $t_j \in S$ appears in the demonstration inputs, it will be replaced by a uniformly randomly picked token $t' \in S$, i.e., $c(t_j) \sim \texttt{uniform}(S)$.
For example, if the token ``school'' appears twice in the demonstration inputs, then they will likely be replaced by two different tokens.
In contrast, in \textsc{Bijective} cipher (\S\ref{subsec:bijectiveciphers}) we ensure multiple occurences of a token are conistently replaced by the same token.

\paragraph{Indecipherability of \textsc{Non-Bijective} cipher:}
In a \textsc{Non-Bijective} cipher, the mapping is no longer reversible, which means it's impossible for models to learn the mapping nor recover the original inputs.
This is because $c$ is not surjective anymore, and hence $c^{-1}$ does not exist.
This implies that a composite function through which $h$ can be recovered also does not exist.

\subsection{Measuring ``Learning'' via \name{}}
\label{subsec:learning:measuring}

Bijective ciphers offer a novel and challenging yet solvable task encoding, making it unlikely to be seen from pretraining. However, the performance of LLMs on this cipher might be influenced by unciphered tokens ($t \in V \setminus S$), which may invoke task retrieval capability of LLMs.  
In contrast, we quantify ICL `learning'' using \textbf{the performance gap} between \textsc{Bijective} (\S\ref{subsec:iclciphers}) and \textsc{Non-Bijective} (\S\ref{subsec:non-bijectiveciphers}) ciphers.
The comparison between these two ciphers is meaningful because the ciphers always share the same ciphered tokens for consistency. The only difference between the two is their token mapping functions: \textsc{Bijective} cipher mapping allows a reversible mapping of ciphered tokens. In contrast, \textsc{Non-Bijective} cipher removes the learnable patterns. 
Therefore, the gap between the performance on \textsc{Bijective} and \textsc{Non-Bijective} ciphered text can be a practical measure of TL.

Although it's theoretically possible to completely solve the one-to-one mappings of \textsc{Bijective} ciphers, the models are not necessarily required to do so to solve the reformulated tasks. 
Instead, they only need to (internally) capture related information or attributes (e.g. sentiment ) of the ciphered tokens, depending on the tasks and given demonstrations.
Our experiment results in \ref{sec:experiment} show that solving the cipher partially can still help the model better solve the reformulated tasks. 

\section{Experimental Setup}
\label{sec:setup}
We discuss our setup for evaluating 
\name{} when applied to various tasks. 

\subsection{Design Choices for \name}
\label{subsec:empirical}

\paragraph{Zipfian shuffling:}
Literature has shown a strong correlation between token frequency in the pre-training corpus and model performance \citep{razeghi2022impact,mallen2022trustlmretrieval}---LLMs tend to perform better on frequent tokens.
To diminish the confounding influence of token frequency, we constrain the shuffling between tokens of similar frequency. 
Inspired by Zipfian shuffling \cite{piantadosi2014zipf},
 we divide all the tokens into $k$ ($k=10$ in our experiments) groups of similar frequency and shuffle the tokens within each group. 
 Specifically, we use the Wikipedia~\cite{wikidump}  to calculate token frequency instead, which approximates the actual token frequency.
   
\paragraph{Priority sampling of ICL demos:}
\label{subsec:sampling}
To create an ICL demo set, one way is to randomly sample the required number of examples (say $n$) from the pool of demos. We call this \textbf{non-priority} (random) sampling. 
However, in practice we always perform \textbf{priority sampling} (unless otherwise specified) where we prioritize examples that contain 
the substituted tokens of the test case input. 
This is done to expose LLMs to the relevant substitutions from which they can learn to decipher. 
Suppose the number of ciphered tokens in the test input is $m$ (which depends on the shuffle rate $r$).
The goal is to select $n$ demonstrations from the pool of demos, such that each of them contains at least one of the $m$ uniquely ciphered (substituted) tokens. 
This is trivial if $m=n$ (i.e., $n$ demos cover the whole set of $m$ substitutions). Otherwise:

\begin{itemize}[noitemsep,leftmargin=14pt,topsep=3pt]
 \item   If $m < n$ (i.e., the number of substitutions is less than the required number of ICL demos to be sampled from the pool), we choose $m$ examples according to priority sampling and the rest of $n-m$ examples are randomly picked from the demo pool. 
 \item  If $m > n$, we select a random subset of the ciphered tokens of size $n$.
 
\end{itemize}

In \S\ref{appendix:priority-sampling}, we compare priority sampling with non-priority (random) sampling.


\paragraph{Shuffle Rate:} 

The shuffle rate \(r\) determines the proportion of tokens that are substituted.  
When \(r\) is close to 0, the cipher's effect is minimal, as few or no tokens are substituted, making it uninteresting. 
Conversely, when \(r\) approaches 1, nearly all tokens are shuffled and solving the task is almost impossible (under both \textsc{Bijective} and \textsc{Non-bijective} ciphers).  
Thus, our focus lies on a moderate shuffle rate between 0 and 1, striking a balance between these extremes.  
We analyze this in \S\ref{subse:shuffle-rate}.

\begin{table*}[ht]
    \centering
    \small
\begin{tabular}{
P{3cm}@{\hskip 0.2cm}
P{3cm}@{\hskip 0.2cm}
L{2.0cm}@{\hskip 0.2cm}
L{2.0cm}@{\hskip 0.2cm}
L{2.0cm}@{\hskip 0.2cm}
L{2.0cm}@{\hskip 0.2cm}
}
    \toprule
    Model $\rightarrow$ &\multirow{2}{*}{Cipher}& \multicolumn{4}{c}{ 20-shot  } \\
     \cmidrule(lr){3-6}
     Dataset (shuffle rate) $\downarrow$ & & Llama3.1 & Qwen2.5 & Olmo & Gemma2\\
     \midrule
     \rule{0pt}{2ex}
    \multirow{2}{*}{SST-2 ($r=0.5$)} & \textsc{Non-Bijective} & 58.3  & 69.0  & 67.7   & 70.5  \\
    & \textsc{Bijective} & 63.1   {(\textcolor{DarkGreen}{+4.8}  $\uparrow$)}$^*$ & 73.5   {(\textcolor{DarkGreen}{+4.5} $\uparrow$)}$^*$ & 72.7    {(\textcolor{DarkGreen}{+5.0} $\uparrow$)}$^*$ & 74.2   {(\textcolor{DarkGreen}{+3.7} $\uparrow$)}$^*$   \\
    \midrule
    
    \multirow{2}{*}{Amazon ($r=0.6$)} & \textsc{Non-Bijective} & 64.7  & 72.6  & 77.2  & 80.8     \\ 
    & \textsc{Bijective} & 72.3   {(\textcolor{DarkGreen}{+7.6} $\uparrow$)}$^*$ & 77.9   {(\textcolor{DarkGreen}{+5.3} $\uparrow$)}$^*$ & 80.2 {(\textcolor{DarkGreen}{+3.0} $\uparrow$)}$^*$ & 85.0   {(\textcolor{DarkGreen}{+4.2} $\uparrow$)}$^*$ \\
    \midrule
    
    \multirow{2}{*}{HellaSwag ($r=0.3$)} & \textsc{Non-Bijective} & 29.7  & 52.8 & 25.9  & 37.1    \\
    & \textsc{Bijective} & 31.9  {(\textcolor{DarkGreen}{+2.2} $\uparrow$)}$^*$ & 62.3  {(\textcolor{DarkGreen}{+9.5} $\uparrow$)}$^*$ & 26.1   {(\textcolor{DarkGreen}{+0.2} $\uparrow$)}$^*$ & 36.6   {(\textcolor{DarkRed}{-0.5} $\downarrow$)} \\
    \midrule
    
    \multirow{2}{*}{WinoGrande ($r=0.1$)} & \textsc{Non-Bijective} & 53.7  & 61.3  & 53.4  & 63.5    \\
    & \textsc{Bijective} & 55.5   {(\textcolor{DarkGreen}{+1.8} $\uparrow$)}$^*$ & 62.5   {(\textcolor{DarkGreen}{+1.2} $\uparrow$)} & 53.1   {(\textcolor{DarkRed}{-0.3} $\downarrow$)} & 63.5   {(\textcolor{DarkGreen}{+0.0} $\uparrow$)} \\
    \bottomrule
    
    \end{tabular}
    \caption{
    LLM accuracies (reported in \%) with 20-shot demonstrations, under \textsc{Bijective} and \textsc{Non-Bijective} ciphers. 
    For each dataset, we fix the shuffle rate to a reasonable value here to demonstrate the gap. We provide an analysis on the effect of shuffle rate later (\S\ref{subse:shuffle-rate}).
    The numbers inside the parenthesis shows the change from \textsc{Non-Bijective}  to \textsc{Bijective} ciphering (gains in \textcolor{DarkGreen}{green}$\uparrow$ and losses in \textcolor{DarkRed}{red}$\downarrow$). In majority of cases, we observe \textbf{consistent performance gains under \textsc{Bijective} cipher}. 
    Statistically significant gains are indicated via $^*$.
    }
    \label{tab:fixed_shot_r}
\end{table*}

\begin{figure*}[ht]
\centering
\includegraphics[width=0.43\textwidth]{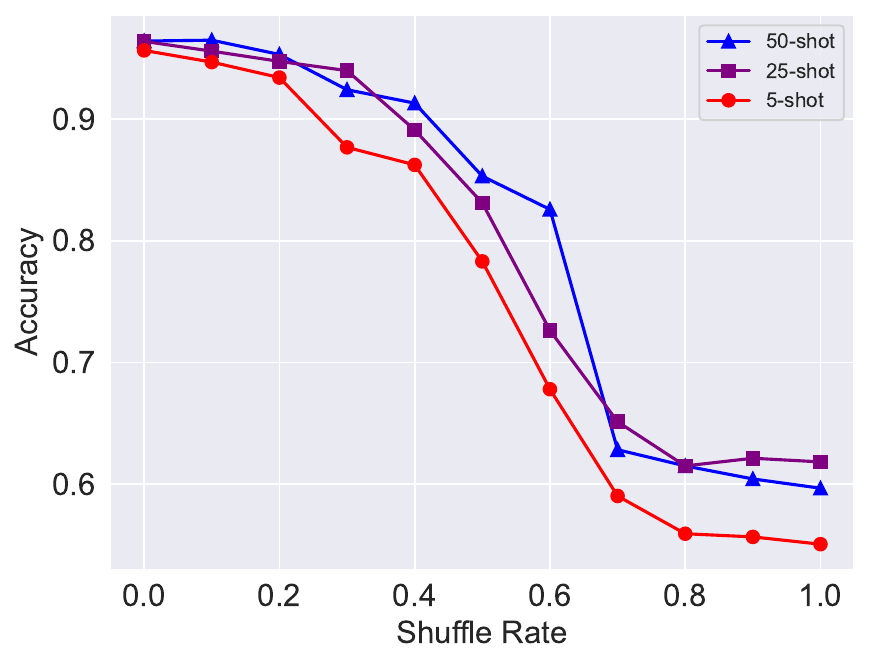}
\includegraphics[width=0.44\textwidth]{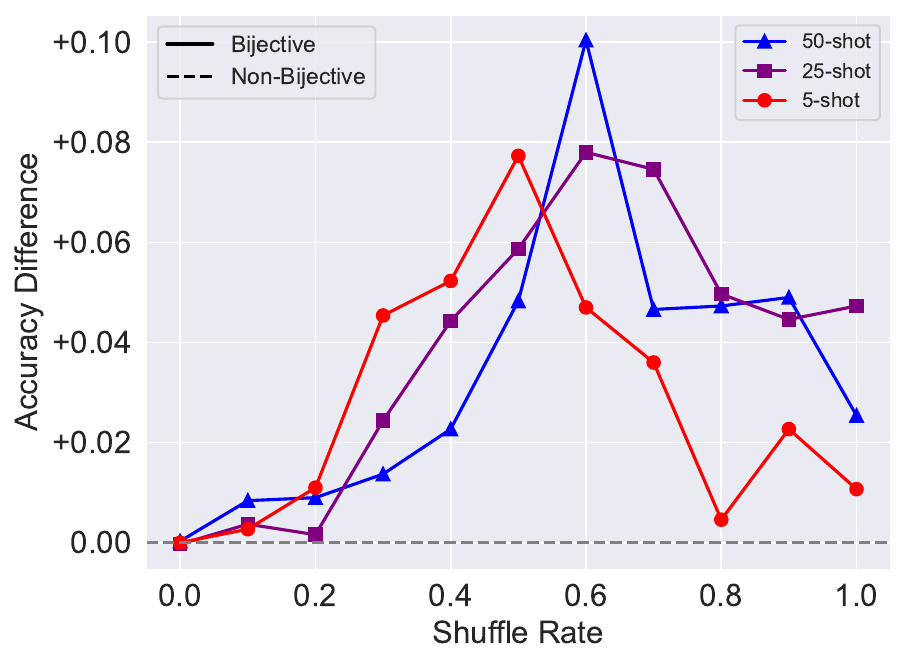}
\caption{
Llama 3.1 8B performance on Amazon dataset.
\textbf{Left:} 
Under the \textsc{Bijective} cipher, accuracy decreases smoothly as the shuffle rate increases, highlighting the difficulty in interpreting the ciphered text. 
Accuracy also increases with more demonstrations, suggesting the model's ability to solve \textsc{Bijective} cipher. 
\textbf{Right:} $y$-axis shows the accuracy gap between \textsc{Bijective} and \textsc{Non-Bijective} ciphers. 
For very high shuffle rates (e.g, $> 0.7$), the task become very hard to understand and solve (for the model and even humans) as it becomes ill-defined.
} \label{fig:amazon-curve}
\end{figure*}

\paragraph{Special tokens and filters:}
LLMs usually have a list of special tokens that help the model understand the prompt and task (e.g. next token prediction). 
For example, Llama3.1 models use  \texttt{<|begin\_of\_text|>} and \texttt{<|end\_of\_text|>} to denote the start of input and end of generation.
We preserve special and punctuation tokens from getting ciphered to avoid hurting models' basic functionality. (Full list of preserved tokens is in Appendix~\ref{appendix:reserved}). Similarly, we avoid disturbing spaces in the original text (details in Appendix~\ref{appendix:space}). 


\subsection{Evaluated Models}
\label{subsec:models}
We mainly focus on pretrained LLMs in our experiments, including Llama 3.1 \cite[\texttt{Llama-3.1-8B}]{dubey2024llama3herdmodels}, \qwen \cite[\texttt{Qwen2.5-7B}]{qwen2.5}, \olmo \cite[\texttt{OLMo-7B-0724-hf}]{Groeneveld2023OLMo} and \gemma \cite[\texttt{Gemma-2-9b}]{gemma_2024}. 
In \S\ref{subsec:instruct} and \S\ref{subsec:size}, we also show results on {\tt Llama-3.1-8B-Instruct} and {\tt Llama-3.1-70B} to explore the effect of instruction tuning and model size. 
Unless otherwise specified, 
\llama~refers to {\tt Llama-3.1-8B}.

\subsection{Datasets}

\label{subsec:datasets}
We conduct experiments on four datasets. 
SST-2 \cite{socher2013recursive} and Amazon \citep[Amazon Reviews 2023]{hou2024bridging} are binary sentiment classification tasks. 
HellaSwag \cite{zellers2019hellaswag} is a sentence completion task, formatted as four-choices QAs.
WinoGrande \cite{sakaguchi2020winogrande} is a pronoun resolution task, formatted as binary-choice QAs.
For each dataset, we curate a demo pool for sampling ICL demos, and a test set contains to-be-tested cases.
We use accuracy as the metric for all our experiments if not specified.
We averaged the metrics across three runs of experiments for a more reliable evaluation.
Further details on datasets (prompts and examples) are in \S\ref{appendix:additional details} and \S\ref{appendix:examples}.

\section{Empirical Findings}
\label{sec:experiment}
We evaluate \name{} on a range of LLMs and datasets. 
We then use the accuracy gap between the two types of ciphers to quantify a proxy for TL capabilities of LLMs (\S\ref{subsec:learning:measuring}). 
\subsection{Evidence of Task-Learning in ICL}
\label{results:llm:solve:it}
\label{results}

\autoref{tab:fixed_shot_r} shows the performance of LLMs on four datasets ciphered with our framework (\S\ref{sec:framework}), with a fixed shuffle rate and number of demonstrations. 
The statistically significant results are marked with $^*$ using McNemar's test \cite{McNemar_1947}. 
The null hypothesis is that two marginal probabilities for each outcome are the same, meaning switching from \textsc{Non-bijective} to \textsc{Bijective} cipher has no impact on the prediction results.
We see a consistent improvement in the performance of LLMs on \textsc{Bijective} ciphered inputs over \textsc{Non-bijective} ciphered inputs (except for Olmo on WinoGrande and \gemma on Hellaswag). 
This consistent gap demonstrates that \textbf{LLMs solve decipherable \textsc{Bijective} ciphers better than the undecipherable \textsc{Non-bijective} ones}. 
This provides evidence for 
\textbf{task learning} capabilities of LLMs. 



\subsection{Analysis: Effect of Shuffle Rates}
\label{subse:shuffle-rate}

As discussed in \S\ref{subsec:empirical}, the shuffle rate \(r\) dictates the percentage of tokens that are substituted. When \(r\) is near 0, the cipher has little to no impact. When \(r\) nears 1, almost all tokens are shuffled, making the task nearly unsolvable. 
Therefore, we expect the largest difference between \textsc{Bijective} and \textsc{Non-bijective} ciphers when \(r\) is somewhere between the two extremes.
We verify this intuition in 
Fig.\ref{fig:amazon-curve} which shows the performance of \llama~on the Amazon dataset with priority sampling. 
We can observe the largest gap between \textsc{Bijective} and \textsc{Non-bijective} ciphers across the interval $r \in (0.4, 0.6)$, which aligns with expectations. 

\subsection{Analysis: Effect of Number of Demos}
Prior work~\citep{srivastava2023beyond} shows ICL performance improves with more demonstrations.
\autoref{tab:llama8b_infor_sample} reports performance gaps between \textsc{Bijective} and \textsc{Non-bijective} ciphers as the number of demos varies.
\textsc{Bijective} consistently outperforms \textsc{Non-bijective}, with the gap widening as demos increase—though this effect plateaus beyond a point, particularly for Hellaswag and WinoGrande.
Fig.\ref{fig:amazon-curve} (on the right) also shows this visually for the Amazon dataset. 

{
\setlength{\tabcolsep}{0pt}
\begin{table*}[ht]
    \centering
    \small
\resizebox{\linewidth}{!}{
\begin{tabular}{
P{2.8cm}@{\hskip 0.2cm}
P{2.5cm}@{\hskip 0.2cm}
L{1.85cm}@{\hskip 0.2cm}
L{1.85cm}@{\hskip 0.2cm}
L{1.85cm}@{\hskip 0.2cm}
L{1.85cm}@{\hskip 0.2cm}
L{1.85cm}@{\hskip 0.2cm}
L{1.95cm}@{\hskip 0.2cm}
}
    \toprule
    Shots $\rightarrow$ &\multirow{2}{*}{Cipher}& \multicolumn{6}{c}{Model: Llama 3.1 8B} \\
     \cmidrule(lr){3-8}
     Dataset (shuffle rate)$\downarrow$ & & 5-shot & 10-shot & 15-shot & 20-shot & 25-shot & 50-shot \\
     \midrule
     \multirow{2}{*}{SST-2 ($r=0.5$)} & \textsc{Non-Bijective} & 56.9  & 59.5  & 58.6  & 58.3   & 62.6  & 58.4    \\
     & \textsc{Bijective} & 59.5  {(\textcolor{DarkGreen}{+2.6} $\uparrow$)}$^*$ & 61.0  {(\textcolor{DarkGreen}{+1.5} $\uparrow$)} & 60.8  {(\textcolor{DarkGreen}{+2.2} $\uparrow$)} & 63.1  {(\textcolor{DarkGreen}{+4.8} $\uparrow$)}$^*$ & 65.4  {(\textcolor{DarkGreen}{+2.8} $\uparrow$)}$^*$ & 64.9  {(\textcolor{DarkGreen}{+6.5} $\uparrow$)}$^*$ \\
    \midrule
    \multirow{2}{*}{Amazon ($r=0.6$)} & \textsc{Non-Bijective} & 63.1 & 61.8  & 68.1  & 64.7  & 64.8  & 72.5   \\
    & \textsc{Bijective} & 67.8  {(\textcolor{DarkGreen}{+4.7} $\uparrow$)}$^*$ & 67.6  {(\textcolor{DarkGreen}{+5.8} $\uparrow$)}$^*$ & 74.5  {(\textcolor{DarkGreen}{+6.4} $\uparrow$)}$^*$ & 72.3  {(\textcolor{DarkGreen}{+7.6} $\uparrow$)}$^*$ & 72.6  {(\textcolor{DarkGreen}{+7.8} $\uparrow$)}$^*$ & 82.6  {(\textcolor{DarkGreen}{+10.1} $\uparrow$)}$^*$ \\
    \midrule
    
    \multirow{2}{*}{HellaSwag ($r=0.3$)} & \textsc{Non-Bijective} & 31.7  & 29.7   & 30.7   & 29.7   & 30.9   & 33.1     \\
    & \textsc{Bijective} & 34.2     {(\textcolor{DarkGreen}{+2.5} $\uparrow$)}$^*$ & 31.7   {(\textcolor{DarkGreen}{+2.0} $\uparrow$)} & 34.1  {(\textcolor{DarkGreen}{+3.4} $\uparrow$)}$^*$ & 31.9    {(\textcolor{DarkGreen}{+2.2} $\uparrow$)}$^*$ & 31.6   {(\textcolor{DarkGreen}{+0.7} $\uparrow$)} & 33.9   {(\textcolor{DarkGreen}{+0.8} $\uparrow$)} \\
    \midrule
    
    \multirow{2}{*}{WinoGrande ($r=0.1$)} & \textsc{Non-Bijective} & 54.9   & 53.2    & 53.7    & 53.7    & 53.3    & 54.3   \\
    & \textsc{Bijective} & 56.3   {(\textcolor{DarkGreen}{+1.4} $\uparrow$)} & 53.8  {(\textcolor{DarkGreen}{+0.6} $\uparrow$)}$^*$ & 54.2  {(\textcolor{DarkGreen}{+0.5} $\uparrow$)}$^*$ & 55.5  {(\textcolor{DarkGreen}{+1.8} $\uparrow$)}$^*$ & 54.6  {(\textcolor{DarkGreen}{+1.3} $\uparrow$)}$^*$ & 55.5  {(\textcolor{DarkGreen}{+1.2} $\uparrow$)}$^*$ \\
    \bottomrule
    \end{tabular}
    }
    \caption{Llama3.1 8B accuracies (reported in \%) on different datasets with varying numbers of ICL examples under \textsc{Bijective} vs. \textsc{Non-Bijective} ciphers. 
    The numbers inside the parenthesis shows the change from \textsc{Non-Bijective}  to \textsc{Bijective} cipher. 
    Statistically significant gains are indicated via $^*$. 
    }
    \label{tab:llama8b_infor_sample}
\end{table*}
}

\subsection{Analysis: Effect of Alignment}
\label{subsec:instruct}
\begin{figure*}[h]
\centering
\includegraphics[width=\textwidth]{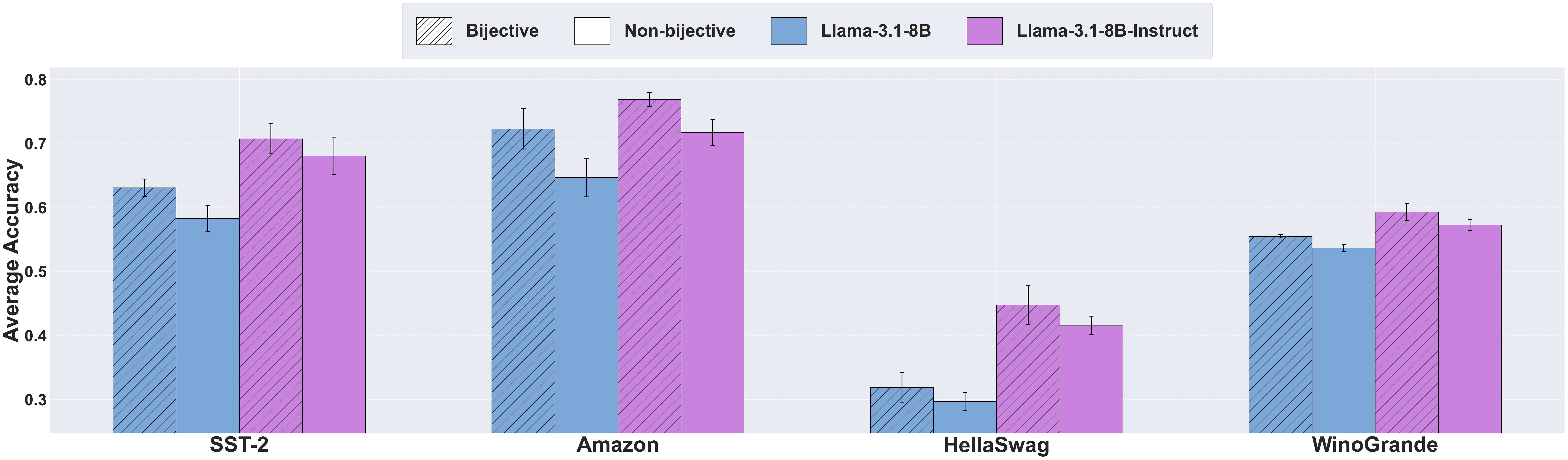}
\caption{Accuracy comparison of Llama-3.1-8B and Llama-3.1-8B-Instruct on four datasets under \textsc{Bijective} and \textsc{Non-bijective} ciphers with 20-shot (\S\ref{subsec:instruct}). 
\textbf{Both aligned and non-aligned models achieve similar relative improvements when solving tasks encoded with a \textsc{Bijective} cipher, compared to those encoded with \textsc{Non-Bijective} ciphers}. 
}
\label{fig:model_ins}
\end{figure*}

Thus far, we have shown results on pre-trained models (before alignment). Here we verify if the results hold up on aligned (e.g., instruction-tuned) models. 
Fig.\ref{fig:model_ins} compares {\tt Llama-3.1-8B} (not aligned) and {\tt Llama-3.1-8B-Instruct} (aligned), which quantifies the effect of alignment on the gaps between \textsc{Bijective} and \textsc{Non-bijective} ciphers.
As expected, the aligned model ({\tt Llama-3.1-8B-Instruct}) outperforms the non-aligned model  ({\tt Llama-3.1-8B}), on both \textsc{Bijective} and \textsc{Non-bijective} ciphers. 
But crucially, the gaps between the two ciphers remain similar in both settings. 
This indicates that \textbf{the decipherability for \textsc{Bijective} ciphers is maintained in aligned models}.
\S\ref{appendix:ins-model} shows more complete results on Llama3.1-8B-Instruct.

\subsection{Analysis: Effect of Model Size}
\label{subsec:size}
\begin{figure*}[h]
\centering
\includegraphics[width=\textwidth]{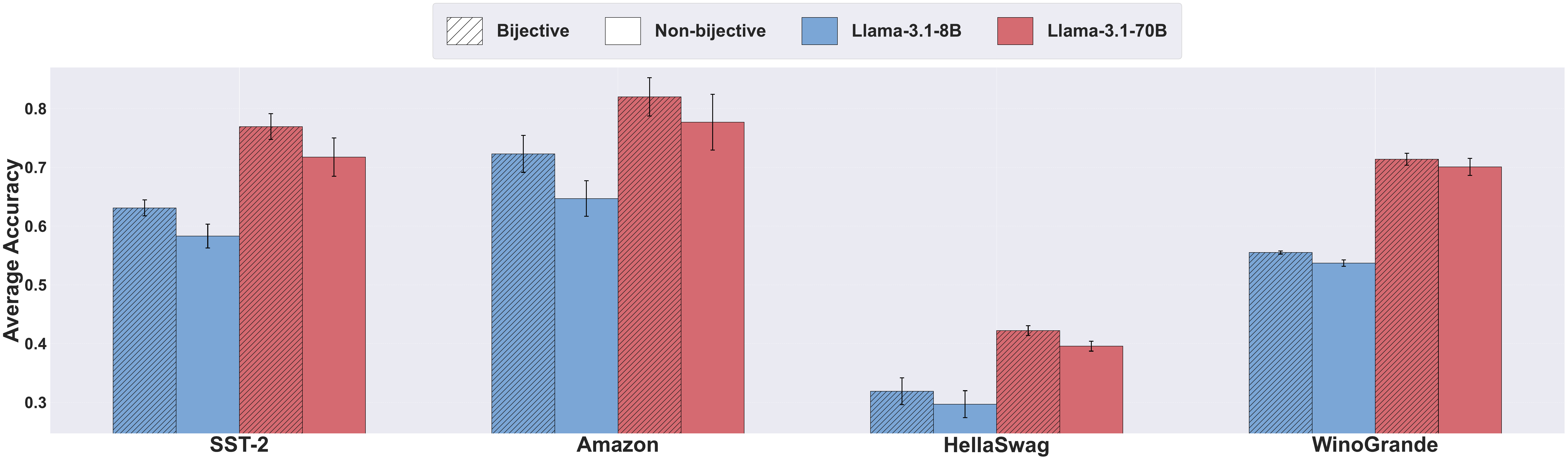}
\caption{Accuracy comparison of Llama-3.1-8B and Llama-3.1-70B on four datasets under \textsc{Bijective} and \textsc{Non-bijective} ciphers with 20-shot (\S\ref{subsec:size}). 
Larger models outperform smaller ones under both ciphers, while \textsc{Bijective} cipher consistently yields higher accuracy than \textsc{Non-Bijective} cipher. 
}
\label{fig:model_size}
\end{figure*}

Fig.\ref{fig:model_size} compares {\tt Llama-3.1-8B} and {\tt Llama-3.1-70B}, showing the effect of model size on the gaps between \textsc{Bijective} and \textsc{Non-bijective} ciphers. 
As the model size increases, performances for both \textsc{Bijective} and \textsc{Non-bijective} ciphers improve. 
The gaps between the two ciphers remains similar in the large model, indicating the decipherability of \textsc{Bijective} ciphers across models of different sizes.
We do \emph{not} observe any larger gaps in the large model compared to the small model.
\S\ref{appendix:large-model} shows more complete results on {\tt Llama3.1-70B}.

\subsection{Analysis: Effect of Grammatical Roles}
\label{sec:role}

As shown in Fig.\ref{fig:pipeline}, the substitution/ciphering process may happen between tokens of different POS groups, which changes the syntactic and semantic structure of natural language. 
To explore how ciphers affect tokens differently based on their grammatical roles, we restrict the space of vocabulary shuffling to only one POS group - noun, which maintains the original syntactic and semantic structure.
\autoref{tab:llama8b_noun} shows that gaps between \textsc{Bijective} and \textsc{Non-Bijective} ciphers when shuffling within nouns are similar to those when shuffling within all the tokens.
This indicates ciphering within certain grammatical roles is still a solvable task for the models.

\begin{table}[h]
    \centering
    \small
    \resizebox{1.0\linewidth}{!}{
\begin{tabular}{
P{3cm}@{\hskip 0.2cm}
P{2.2cm}@{\hskip 0.2cm}
L{1.7cm}@{\hskip 0.2cm}
L{1.7cm}@{\hskip 0.2cm}
}
    \toprule
     &\multirow{2}{*}{Cipher}& \multicolumn{2}{c}{ Llama3.1 8B 20-shot  } \\
     \cmidrule(lr){3-4}
     Dataset (shuffle rate) & & All & Noun\\
     \midrule
    
    \multirow{2}{*}{HellaSwag (r = 0.3)} & \textsc{Non-Bijective} & 29.7   & 32.1      \\
    & \textsc{Bijective} & 31.9     {(\textcolor{DarkGreen}{+2.2} $\uparrow$)} & 33.6  {(\textcolor{DarkGreen}{+1.5} $\uparrow$)}   \\
    \midrule
    
    \multirow{2}{*}{WinoGrande (r = 0.1)} & \textsc{Non-Bijective} & 53.7   & 54.3     \\
    & \textsc{Bijective} & 55.5   {(\textcolor{DarkGreen}{+1.8} $\uparrow$)} & 56.7   {(\textcolor{DarkGreen}{+2.4} $\uparrow$)} \\
    \bottomrule
    
    \end{tabular}
    }
    \caption{Llama3.1 8B accuracies (reported in \%) with 20-shot demonstrations, under \textsc{Bijective} and \textsc{Non-Bijective} ciphers. 
    ``All'' operates shuffling on all the tokens while ``Noun'' constrains shuffling to only nouns.
    }
    \label{tab:llama8b_noun}
\end{table}

\subsection{Analysis: Probing Representations}
\label{sec:probing}

\begin{figure*}[h]
\centering
\includegraphics[width=0.45\textwidth]{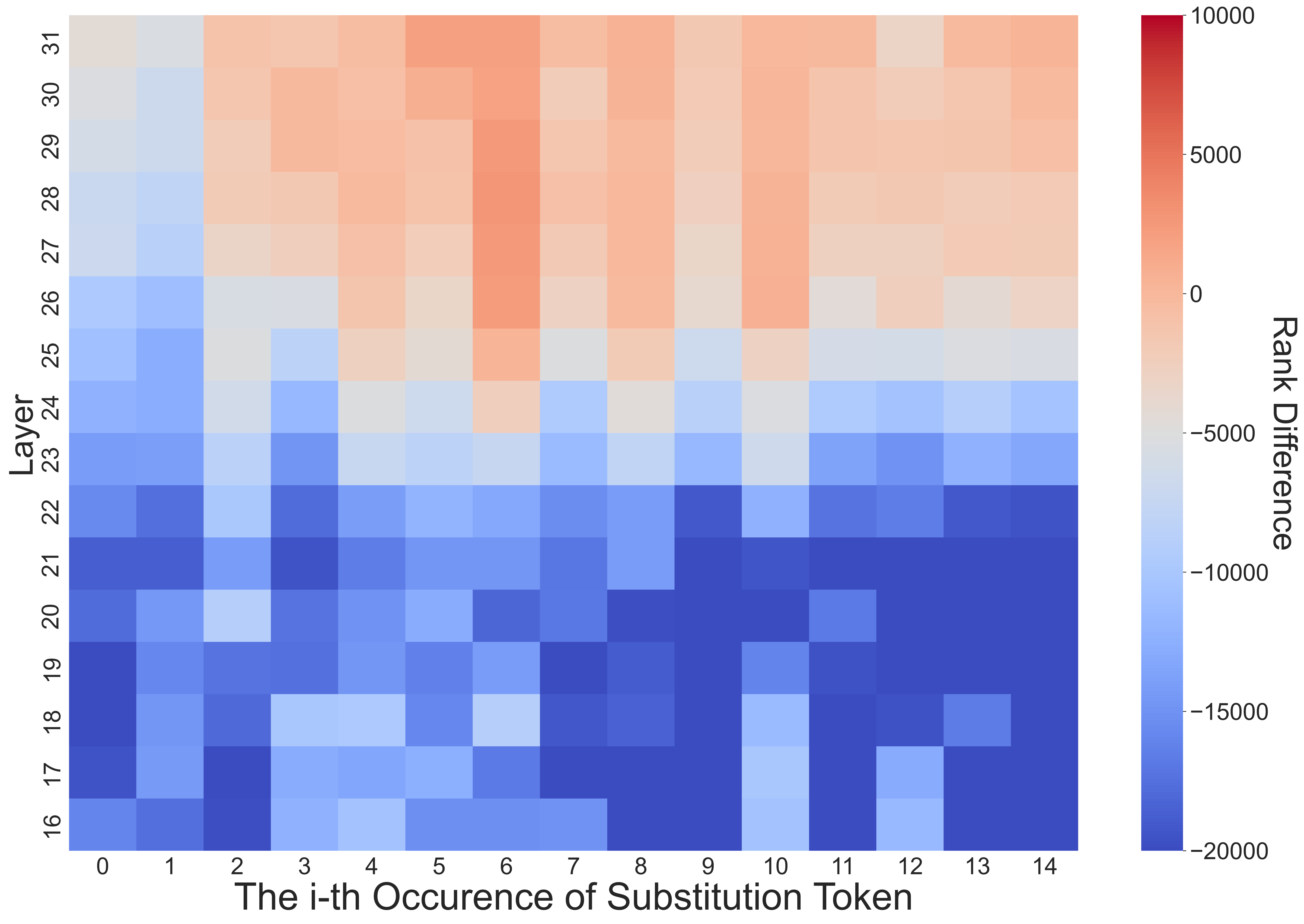}
\hfill
\includegraphics[width=0.45\textwidth]{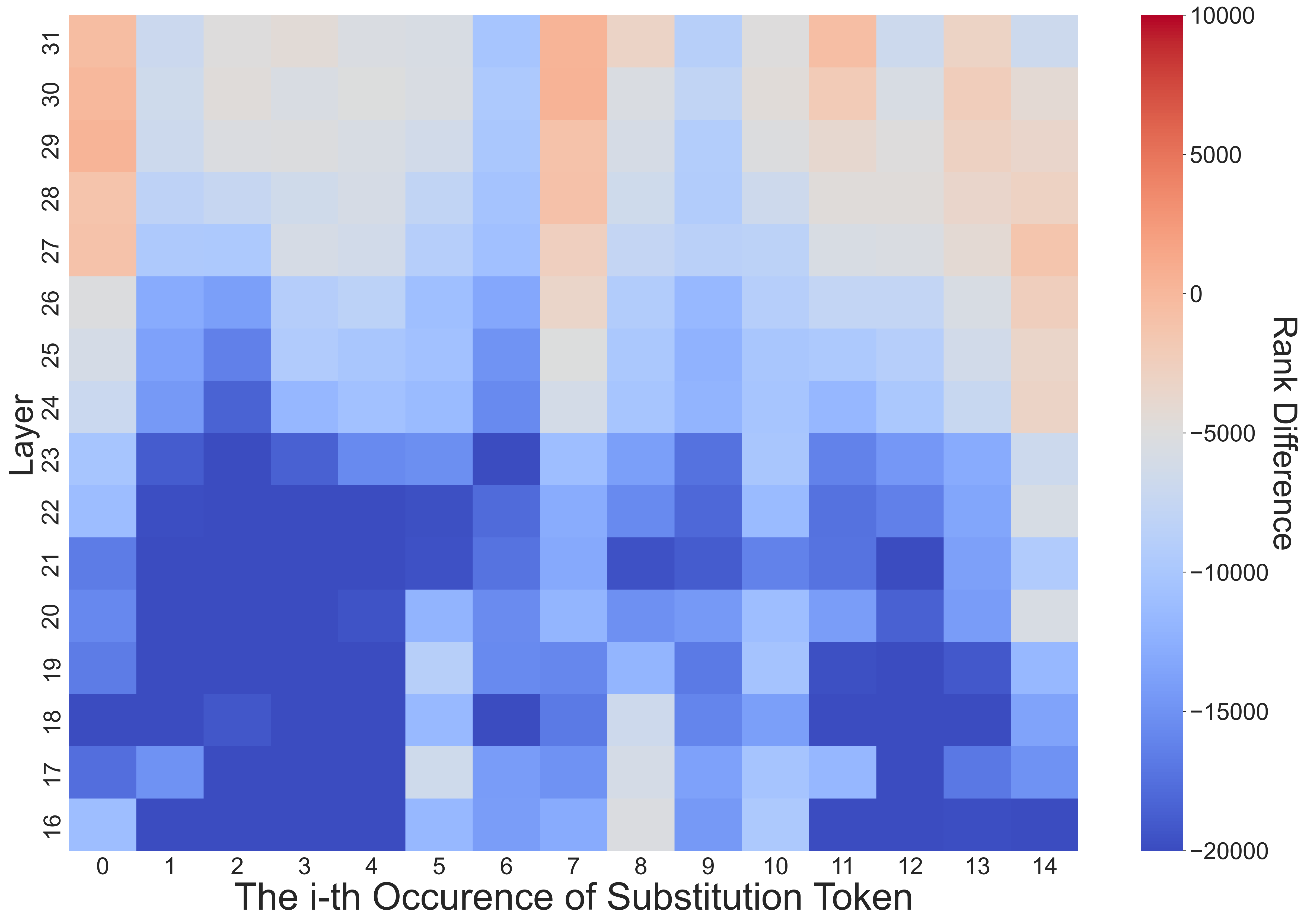}
\caption{
$x$-axis indicates the $i$-th occurrence of ciphered tokens in the \llama{} context.
$y$-axis indicates the rank difference 
(Eq~\ref{rank-diff}). 
Positive values (red) indicate the model's preference for substituted tokens over original ones. In the \textsc{Bijective} cipher (left), we see a preference that favors substituted tokens. However, there is no clear preference in the \textsc{Non-Bijective} cipher (right).    
}
\label{fig:heatmap_half}
\end{figure*}







To examine how LLMs process ciphered inputs, we use Logit Lens \cite{logitlens} to probe their intermediate layer representations. Logit Lens takes token embeddings from intermediate layers and decodes them using the final LM head. We conduct this probing on the Amazon sentiment dataset using \llama 8B. 


\paragraph{Selecting tokens for probing:}
We first pick 600 most frequent tokens in the demo set after filtering out tokens other than verbs, nouns and adjectives, using NLTK~\cite{bird2009natural}.
We randomly sample 30 tokens from them as the ``original tokens''. 
We then randomly sample another 30 tokens from the remaining 570 tokens as the ``substituted tokens'', each of which has a one-to-one correspondence with the original tokens.
\noindent
\textbf{Token substitution:}
For \textsc{Bijective} cipher, we create a bijection between the 30 original tokens and the selected 30 substitution tokens, creating a mapping for the original tokens to be substituted. 
For \textsc{Non-Bijective} cipher, we substitute each occurrence of each original token, by a randomly sampled token from the remaining 570 tokens.

\noindent
\textbf{Building ciphered inputs:}
For each original token $t'$ (the token to be ciphered), we sample 15 examples from the demo pool that contain $t'$, and apply our two substitution ciphers to build the ciphered prompt. 
Given the positions of original tokens $P = (p_1, p_2, ..., p_n)$, we apply the Logit Lens and observe embeddings at positions $P' = (p_1-1, p_2-1, ..., p_n-1)$ (i.e., one position prior) to find the ranks of original tokens and ``substituted tokens''. This gives us an understanding of how the model changes its preference between original and substituted tokens. 
We quantify this notion as the rank difference (original token rank - substitution token rank):  
\begin{equation}
\text{rank-diff} = \text{rank}(t_j) - \text{rank}(c(t_j)),
\label{rank-diff}
\end{equation}
where rank denotes the position of a given token in the model's softmax score over the vocabulary set.

\noindent
\textbf{LLM representations favor substituted tokens in \textsc{Bijective} cipher:}
For \textsc{Bijective} cipher (Fig.\ref{fig:heatmap_half}; left) as the model observes more substitutions, the rank difference changes from negative to positive (in deeper layers, where the model interpreting with LogitLens is more meaningful). 
Consistently, the model gives a higher rank to the substituted tokens than the original tokens, suggesting that the model starts to understand the cipher. 
In contrast, there is no trend for \textsc{Non-Bijective} cipher (Fig.\ref{fig:heatmap_half}; right) as there is nothing to decipher. 




\section{Related Work}
\label{sec:related_work}

\noindent
\textbf{Dual operating modes of ICL:} 
\citet{min2022rethinking} showed the disconnect between ``learning'' and the content of in-context demonstrations (lack of task ``learning''). This motivated following works to identify two primary modes of operation for In-Context Learning (ICL): \emph{task retrieval} (TR), which involves recalling patterns previously encountered in pre-training data, and \emph{task learning} (TL), which involves learning new patterns on-the-fly that were not seen during pre-training. Some studies emphasize TR by exploring the factual recall capabilities of ICL~\citep{sun2022recitation, golchin2024memorization, han2023understanding,zhao2023context,reddy2023mechanistic,dankers2024generalisation}, providing insights into how LLMs memorize pre-training data, thus facilitating TR. 
Other studies ~\cite{lin2024dual,song2024out,vacareanu2024words,nafar2024learning,anand2024dual} have made efforts to measuring ``learning'' in ICL, but focus on simplified datasets (e.g., linear regression) or architectures (e.g., shallow transformers), which differ from our focus. 
Additionally, some of the studies ~\cite{vacareanu2024words,nafar2024learning} may still suffer from data contamination, thus failing to accurately reflect the actual capacity of TL. 
In contrast, our method is aimed at real datasets and real LLMs. 
Most pertinent to our work, 
\citet{pan2023incontext,wang2024investigating} have attempted to separate TR and TL through \emph{output} intervention by replacing labels with abstract symbols like numbers or letters. However, it remains uncertain whether using abstract labels effectively eliminates the influence of TR in ICL. 
Many human-readable tasks may have inherent priors embedded in the pre-training datasets, suggesting that LLMs might still use inputs and prompt structures to infer the task, thereby engaging in implicit task retrieval. 
Our approach proposes an alternative method for quantifying TL by intervening in the \emph{input} space.
Compared to prior works, it is more general as it’s a framework that can be combined with almost all the tasks, and more reliable as it eliminates the effect of data contamination as discussed in Sec~\ref{subsec:learning:measuring}.

\paragraph{Ciphers and their use in AI:} Substitution ciphers are studied in NLP for their potential to decipher lost languages without parallel corpora~\cite[\emph{inter alia}]{knight2006unsupervised, ravi2008attacking, ravi2011bayesian, dou2012large, berg2013unsupervised, pourdamghani2017deciphering, nuhn2013beam, berg2011simple, corlett2010exact, aldarrab2020can}. For example, \citet{ravi2011bayesian} propose a Bayesian approach combining n-gram models and dictionaries for efficient sampling-based decipherment. Deterministic methods also exist, using optimization or heuristics~\cite{peleg1979breaking, ganesan1993statistical, olson2007robust}. \citet{yuan2023gpt} is the only work we know applying ciphers to LLMs (GPT-4) in the context of safety.

\section{Discussion and Conclusion}






\paragraph{\textsc{Bijective} cipher is not a \emph{single} task.} 
The proposed ciphers are a broad reformulation mechanism of \textit{existing} tasks. The underlying task can be \textit{any task} chosen by the user. Our method offers a general-purpose framework for task reformulation that enables us to probe the boundary between memorization and generalization. 
Moreover, the reformulated tasks are different from each other. 
When solving the reformulated tasks, the model doesn't necessarily follow a manner that first completely solve the cipher/mapping, then recover and solve the original tasks. 
Instead, it only needs to (internally) capture related information or attributes (e.g. sentiment ) of the ciphered tokens, depending on the tasks and given demonstrations. 
This means for each reformulated task, the model doesn't always have to completely solve the cipher/mapping and learns differently.

\paragraph{Does \textsc{Bijective} cipher guarantee measuring \textit{only} ``learning''?} 
Achieving a perfect distinction between ``learning'' and ``retrieval'' may be 
unattainable, as any learning inherently involves non-zero level of retrieval (e.g., language understanding). 
Our framework provides a method to quantify learning, by analyzing the difference between how LLMs process a random but learnable bijection, vs non-bijective noise.
Though understanding the complementarity of these approaches and success at quantifying pure learning remains to be further understood in future work. 


\noindent
\textbf{Do the modest gains of \textsc{Bijective} cipher indicate that the weakness of  ``learning'' in ICL?} 
Not necessarily. The proposed re-encoding of ICL transforms tasks into more complex problems that are inherently more challenging to solve. 
This is a feature, not a bug, as it allows us to argue that such esoteric encoding tasks reduce the potential confounding effect of retrieval. However, the side effect is that this increased difficulty in task re-encoding results in smaller gains. The key point is that there are consistent positive gains between the \textsc{Bijective} and \textsc{Non-Bijective} settings. The magnitude of this gap is a secondary consideration and is likely to change with future innovative methods for re-encoding tasks.

\paragraph{Can your results be due to data contamination?}
Our work is motivated by the same issue.
Data contamination makes it difficult to attribute the success of ICL to ``retrieval'' (from pre-training) vs ``learning'' (from in-context demonstrations, without seeing them a priori). A reasonable approach to measure the latter (and mitigate the former) is through randomized tasks. The point of our study is to substitute the given tasks with randomly generated bijection tokens, which makes it impossible for any model to have memorized them. We report the \textit{difference} in performance with bijective vs non-bijective ciphering and de-emphasize any absolute performance numbers which could have resulted from memorization of the original task. 

\paragraph{To measure TL, why don't we just evaluate LLMs on ``novel'' tasks?}
There is currently no straightforward way to define task ``novelty''. Prior work has shown that LLM performance correlates strongly with the presence of tasks in the pretraining data~\citep{razeghi2022impact,mallen2022trustlmretrieval}. To quantify novelty, one would need to either (i) perform large-scale fuzzy matching against pretraining corpora, or (ii) recast tasks into an equivalent representation that is unlikely to have been encountered during pretraining. Few works have tried (i) and have shown some success, but we also know that it's brittle and challenging. Hence, our work focuses on (ii).

\paragraph{Conclusion:}
We introduced \name, a class of cryptography text transformations designed to evaluate novel task learning capabilities of LLMs. 
We show that LLMs exhibit the capacity to decipher these novel tasks during inference. This evidence indicates LLMs' ability to learn novel tasks outside of their pre-training corpus. 
The exact mechanism of this ``learning'' remains an active area of study. 
Understanding this mechanism holds the potential to
unleash novel problem-solving capabilities of LLMs. 
\section*{Limitations}
We discuss the potential limitations of our work: 


\paragraph{Deviation from natural language:}
Ciphered text generated 
diverges from natural language. While this is useful to assess LLMs' TL capabilities, it may also make the task excessively challenging for them. 
Except for restricting the space of shuffling (Sec \ref{sec:role}), it is possible there might be alternative ways to measure learning while maintaining the naturalness of the tasks. 


\paragraph{More models and datasets:}
Although we evaluated 24 settings (six models $\times$ four datasets), expanding our study to include more and larger models would strengthen our findings. 
The largest model we tested was Llama 3.1 70B, due to limited computing resources. 
Additionally, we did not evaluate large, aligned models such as GPT-4-o1, or Gemini. 
Anecdotal evidence suggests that aligned models may lose their ability to follow in-context demonstrations~\cite{fu2022does}, a crucial aspect of our task definition. 
However, we acknowledge that our task could potentially be adapted into a task description or instruction format suitable for aligned models, which deviates from our current setting and could be explored in future work. It would also be interesting to evaluate \name{} on various pre-training checkpoints to better understand how ICL ``learning'' emerges through pre-training.

\paragraph{Deeper 
interpretability analysis:}
In terms of interpretability analysis, we experimented with several approaches (e.g., PatchScope~\cite{ghandehariounpatchscopes}) but found success only with the simplest method, the Logit Lens. More advanced interpretability analyses could provide deeper insights into the underlying mechanisms, offering a clearer understanding of the processes involved.

We recognize these as areas for further exploration and encourage future research to address these limitations.

\section*{Acknowledgements}
This work is supported by ONR grant (N00014-24-1-2089). 
We sincerely thank Dongwei Jiang, Jack Zhang, Andrew Wang, and Hannah Gonzalez for their constructive feedback on an earlier version of this document.

\bibliography{ref}

\appendix 
\onecolumn

\begin{center}
{\Large \textbf{Supplemental Material}}
\end{center}

\startcontents[appendix]
\printcontents[appendix]{l}{1}{\section*{Appendix Contents}}

\section{Additional Related Work}
\paragraph{Alternative explanations of ICL:} Since the discovery of ICL~\citep{brown2020language}, numerous studies have explored it across various contexts~\citep{zhao2021calibrate,min2022rethinking,mishra2022reframing,han2023understanding,wang2023selfinstruct,sia2024does,vacareanu2024words,mueller2024context}. For example, \citet{perez2021true,Lu2022FantasticallyOP,mishra2022reframing} demonstrated ICL's sensitivity to the selection and sequence of demonstrations, while \citet{shin2022effect,razeghi2022impact} highlighted its sensitivity to the frequency and size of the relevant pre-training corpus. Another research direction seeks to elucidate the mechanisms behind ICL. \citet{xie2021explanation} described ICL as implicit Bayesian inference, where ICL demonstrations are mapped to a latent concept (task) learned during pre-training. Other works have attempted to explain ICL as a form of implicit optimization (gradient descent and its variants)~\citep{garg2022can,zhang2023trained,dai2022can,von2023transformers,li2023closeness}, though the applicability of these formalisms to real LLMs is debated~\citep{shen2024icl_vs_gd}. A few studies aim to understand how ICL emerges in LLMs. \citet{hahn2023theory} suggested that the compositional structure of natural language leads to emergent in-context learning, while other works~\citep{chan2022data} propose that certain distributional properties in the pre-training data may give rise to ICL. 
Although these studies offer varying perspectives into the origin and functioning nature of ICL, we propose to disentangle TL and TR components of ICL by observing LLMs' behavior on randomly generated bijections vs. non-bijection noise.

\paragraph{Empirical understanding of ICL:} Ever since In-Context Learning was discovered~\citep{brown2020language}, multiple works have studied it under diverse settings~\citep{zhao2021calibrate,min2022rethinking,mishra2022reframing,han2023understanding,wang2023selfinstruct,sia2024does,vacareanu2024words,mueller2024context}. 
For instance, \citet{srivastava2023beyond} benchmarked ICL under multiple tasks and models; \citet{perez2021true,Lu2022FantasticallyOP} showed the sensitivity of ICL to the choice of demonstrations and their orderings; \citet{shin2022effect,razeghi2022impact} showed the sensitivity of ICL performance to the frequency and size of the relevant pre-training corpus. 
These works have made useful observations that allow us to better use this elusive quality of LLMs.

\paragraph{Functional nature of ICL:} A more recent line of study aims to understand how ICL actually works in LLMs. 
Multiple works have compared ICL with implicit optimization (specifically gradient descent)~\citep{garg2022can,zhang2023trained,dai2022can,akyurek2022learning,von2023transformers,li2023closeness,kim2024transformers}. This line of work claims that Transformers can meta-learn to perform optimization of internal models given a set of demonstrations. However, their study setup with toy transformers does not align with how LLMs are trained as shown by~\citet{shen2024icl_vs_gd}. Moreover, this line of study does not explain the TR capabilities of LLMs.

\paragraph{Forces that lead to ICL:} Few works try to understand \emph{how ICL emerges in LLMs}. \citet{xie2021explanation} explained ICL as implicit Bayesian inference, which maps a ICL demonstrations to a latent concept (task) learned via pre-training. \citet{hahn2023theory} posited that compositional structure in natural language gives rise to emergent in-context learning. Other works~\citep{chan2022data} theorize more distributional properties in the pre-training data, that may give rise to ICL. Many of these works explain some properties of ICL but fail at others. 
The exact origin of ICL in LLMs still remains an active area of study.

\section{Additional Experimental Details}
\label{appendix:additional details}
\subsection{Preserved Tokens}
\label{appendix:reserved}
For Llama 3.1, we preserve the tokens whose ids range from 0 to 255, 128000 to 128256.
For Qwen 2.5, we preserve the tokens whose ids range from 0 to 255, 151643 to 151664.
For OLMo, we preserve the tokens whose ids range from 0 to 244, 50254 to 50279.
For Gemma 2, we preserve the tokens whose ids range from 0 to 472, 255968 to 255999.
For all the models, we preserve the spaces and underlines to ensure the framework of each task.
For example, in the WinoGrande dataset, LLMs are asked to predict the pronouns in a sentence, where the original pronouns are replaced by an underline. 

\subsection{Handling of White Space}
\label{appendix:space}
LLMs encode the spaces between words differently depending on their tokenization. 
\gemma uses a special underline to represent a space, while \llama, \qwen and \olmo uses 'Ġ'. 
There are usually two versions of the same word -- with or without a space before it, which corresponds to two different tokens.
Take \llama for example, the encoded id of ``is'' is 285 while that of ``Ġis'' is 374.
We name tokens containing a space at the beginning as ``space tokens'' and the others as ``non-space tokens''.
To avoid disturbing spaces in the original text, which may confuse the model, we constrain the shuffling to be within their space/non-space sets.



\subsection{Design choices for \name{}}
In Tab.\ref{tab:design_elements}, we explain our design strategies for choosing priority sampling (in selecting demonstrations from the demo pool) and zipfian shuffling (in choosing the mapping $c$).

\definecolor{lightergray}{RGB}{230,230,230}
\definecolor{DarkRed}{RGB}{130,25,0}
\definecolor{DarkGreen}{RGB}{30,130,30}

\newcommand{\cmark}{\multirow{1}{*}{\textcolor{DarkGreen}{\ding{51}}}}
\newcommand{\xmark}{\multirow{1}{*}{\textcolor{red}{\ding{55}}}}

\begin{table}[ht]
    \small 
    \centering
    \begin{tabular}{>{\raggedright\arraybackslash}p{0.16\textwidth}>{\raggedright\arraybackslash}p{0.40\textwidth}>{\raggedright\arraybackslash}p{0.40\textwidth}}
    \toprule
    \textbf{Strategies for ...} & \textbf{Variant 1} & \textbf{Variant 2} \\
    \midrule
    selecting (sampling) demonstrations & \textbf{Priority}: select demonstrations that contain the target substitution in the test example~\cmark & \textbf{Non-priority}: select demonstrations randomly~\xmark \\
    \midrule
    choosing the token mapping $c$
    & \textbf{Zipfian}:
    $c$ maps tokens of similar frequency (popularity) among each other~\cmark 
    & \textbf{Non-Zipfian}: $c$ maps tokens irrespective of their frequency (popularity)~\xmark \\
    
    \bottomrule
    \end{tabular}
    \caption{Design choices for experiments in \name{} discussed in \S\ref{subsec:empirical}. }
    \label{tab:design_elements}
\end{table}

\subsection{Datasets}
\label{appendix:datasets}

For SST-2, HellaSwag and WinoGrande no label provided for the test set. Therefore, we use their validation set instead.

\paragraph{SST-2:}
We use its validation set as our test set, which has size of 872. 
Its training set, which contains 67.3k examples, is used as the demo pool. 
\paragraph{Amazon:}
To fit the Amazon dataset into binary sentiment classification framework, we filter ratings 4-5 as positive and 1-2 as negative (discard rating 3).  
We focus on reviews under the the ``All\_Beauty'' category in our experiments. We sample 144k positive and negative samples to build the demo pool; and 500 other positive and negative examples as the test set.

\paragraph{HellaSwag:}
We use its validation set as our test set, which contains 444 positive examples and 428 negative examples (872 examples in total). 
Its training set, which contains 38K positive examples and 30k negative examples, is used as the demo pool. 

We randomly sample 1k examples from the validation set as our test set.  
We use its training set as the demo pool, which contains 40k examples.

\paragraph{WinoGrande:}

We use its dev set as the test set, which contains 1267 examples. 
Its xl training set is used as demo pool, which has 40k examples.

\subsection{Prompt Template}
\label{appendix:prompt}
We don't include any instructions in our prompt. 
For SST-2 and Amazon, we use the following prompt template:

\textbf{Input:} \{\textit{input\_demo}\}

\textbf{Output:} \{\textit{label\_demo}\}

...

\textbf{Input: }\{\textit{input\_test}\}

\noindent where \{\textit{input\_demo}\} and \{\textit{label\_demo}\} are the input text and sentiment labels of demonstrations, and \{\textit{input\_test}\} is the input text of test case. 

For HellaSwag and WinoGrande, we use the following prompt template:

\textbf{Question:} \{\textit{question\_demo}\}

\textbf{Options:} \{\textit{options\_demo}\}

\textbf{Answer:} \{\textit{answer\_demo}\}

...

\textbf{Question:} \{\textit{question\_test}\}

\textbf{Options:} \{\textit{options\_test}\}

\noindent where {\textit{question\_demo}\}, {\textit{options\_demo}\} and \{\textit{answer\_demo}\} are the questions, options and correct answers of demos, and {\textit{question\_test}\} and {\textit{options\_test}\} are the question and option of the test case.

\section{Example Inputs/Outputs}
\label{appendix:examples}
Here we display the example inputs/outputs on all the four datasets. Note that in our experiments the original inputs are not included in the prompts.
\begin{tcolorbox}[colback=gray!5!white,colframe=gray!75!black,title=\textbf{
Dataset: SST-2; Model: \qwen; Cipher: \textsc{Bijective}; Shuffle Rate: 0.6
}]
{
\tt
\textbf{Ciphered Input:} been sc Mil Swift the Inch for pen Venezuela Moody  \\
\textbf{Original Input:} been sent back to the tailor for some major alterations\\
\textbf{Output:} negative\\
 \\
\textbf{Ciphered Input:} is born Slovenia of an Platform San sitcom involved also Sr implementedecture embarrassed Swift Malay you reach for the tissues Confederate  \\
\textbf{Original Input:} is born out of an engaging storyline , which also is n't embarrassed to make you reach for the tissues . \\
\textbf{Output:} positive \\
... \\

\textbf{Ciphered Test Input:} allows us Swift hope Esc implementedolan Sr poised Swift cheating a Venezuela career Mr a assembled Kann steak filmmaker Confederate   \\
\textbf{Original Test Input:} allows us to hope that nolan is poised to embark a major career as a commercial yet inventive filmmaker . \\
}
\end{tcolorbox}

\begin{tcolorbox}[colback=gray!5!white,colframe=gray!75!black,title=\textbf{
Dataset: Amazon ; Model: \gemma ; Cipher: \textsc{Bijective}; Shuffle Rate: 0.6 
}]
{
\tt
\textbf{Ciphered Input:} didnSUwell really notice anything mob.  I sink it householder substance Woodward Bean Simple Woodward Senior Caldwell Snowyyn Ato was instance.   \\
\textbf{Original Input:} didn't really notice anything special.  I bought it because of the reviews and the price but honestly, I was disappointed. \\
\textbf{Output:}negative \\
 \\
\textbf{Ciphered Input:}Item arrived regions principle unrest neighbours']modern /><modern urchatosyn Woodward item was calcium steamer principle Counter cap rendering Woodward cover ent since it periodsSUwell Fam Arch anymore Simple iconicBer bottom Simple consequently']modern /><modern urchofficial was wrapped dentist regions principle padded envelope.   \\
\textbf{Original Input:}Item arrived in a quick manner.<br /><br />However, the item was received with a damaged cap rendering the cover useless since it won't snap on anymore and dented bottom and top.<br /><br />It was wrapped tightly in a padded envelope.  \\
\textbf{Output:}negative  \\
... \\

\textbf{Ciphered Test Input:} tried it for cosmetic qualifications perimeter a day spa\u00f2 didnPervers Tehran workil   \\
\textbf{Original Test Input:} tried it for cosmetic procedures in a day spa; didn't really work.  \\
}
\end{tcolorbox}

\begin{tcolorbox}[colback=gray!5!white,colframe=gray!75!black,title=\textbf{
Dataset: Hellaswag; Model: \olmo ; Cipher: \textsc{Bijective}; Shuffle Rate: 0.3
}]
{
\tt
\textbf{Ciphered Question:} Ter Back sits million titled with his Board effective on the keys. the Back    \\
\textbf{Original Question:} A man sits a piano with his hands placed on the keys. the man\\
\textbf{Ciphered Options:} (1) begins playing the titled.\textbackslash n(2) Carlos the keys with million malorn.\textbackslash n(3) beats the titled in million benefitedmic thought.\textbackslash n(4) increases the play for playing.\textbackslash n  \\
\textbf{Original Options:} (1) begins playing the piano.\textbackslash n(2) hits the keys with a mallet.\textbackslash n(3) beats the piano in a rhythmic beat.\textbackslash n(4) increases the volume for playing.\textbackslash n \\
\textbf{Answer:} (1) \\
 ...\\
\textbf{Ciphered Question:} People are noted on the street. million Back   \\
\textbf{Original Question:} People are running on the street. a man \\
\textbf{Ciphered Options:} (1) is wearing poetilts.\textbackslash n(2) limited million drink out Wars million After presidents.\textbackslash n(3) negotiating into million encourages Wars fire.\textbackslash n(4) limited million high jump in million Chris competition.\textbackslash n   \\
\textbf{Original Options:} (1) is wearing stilts.\textbackslash n(2) takes a drink out of a water bottle.\textbackslash n(3) jumps into a pile of fire.\textbackslash n(4) takes a high jump in a bar competition.\textbackslash n   \\
}
\end{tcolorbox}

\begin{tcolorbox}[colback=gray!5!white,colframe=gray!75!black,title=\textbf{
Dataset: WinoGrande ; Model: \llama ; Cipher: \textsc{Bijective}; Shuffle Rate: 0.3 
}]
{
\tt
\textbf{Ciphered Question:} Estonia ferry that my parents story tied I permanent in Johnston permanent Stadium partners bla than my house now because the \_ permanent anchored.   \\
\textbf{Original Question:} The home that my parents had when I was in school was a lot nicer than my house now because the \_ was sophisticated. \\
\textbf{Ciphered Options:} (1) ferry, (2) house   \\
\textbf{Original Options:} (1) home, (2) house \\
\textbf{Answer:}(1) \\
 ...\\
\textbf{Ciphered Question:} Sarah permanent Stadium much better Chart than Maria so \_ always got the easier cases.   \\
\textbf{Original Question:} Sarah was a much better surgeon than Maria so \_ always got the easier cases. \\
\textbf{Ciphered Options:} (1) Sarah, (2) Maria   \\
\textbf{Original Options:} (1) Sarah, (2) Maria   \\
}
\end{tcolorbox}

\clearpage

\section{Priority vs. Non-Priority Sampling}
\label{appendix:priority-sampling}

Fig.\ref{fig:amazon-curve} shows peformance of LLaMa 3.1 8B on Amazon dataset with priority sampling. 
Fig.\ref{fig:sst-curve-random} and Fig.\ref{fig:amazon-curve-random} shows peformance of LLaMa 3.1 8B on SST-2 and Amazon datasets with non-priority sampling. 
Comparing with Fig.\ref{fig:sst-curve} and Fig.\ref{fig:amazon-curve}, they demonstrate similar trends but the performances are more unstable due to the randomness of non-priority sampling. 
Therefore, we use priority sampling throughout our experiments for more steady results.


\begin{figure*}[ht]
\centering
\includegraphics[width=0.46\textwidth,trim=0cm 0.1cm 0cm 0cm]{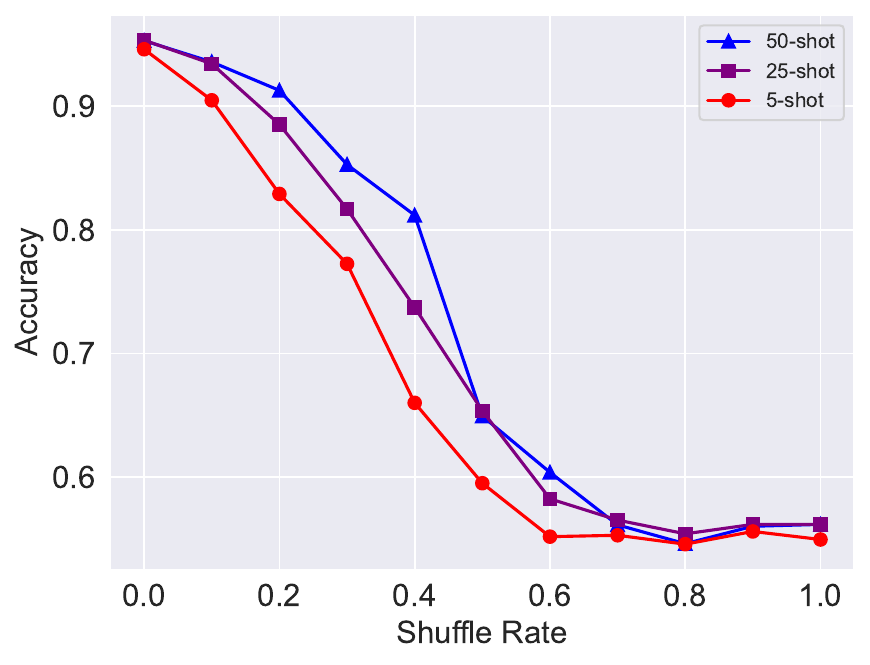}
\hspace{0.5cm}
\includegraphics[width=0.48\textwidth,trim=0cm 0.1cm 0cm 0cm]{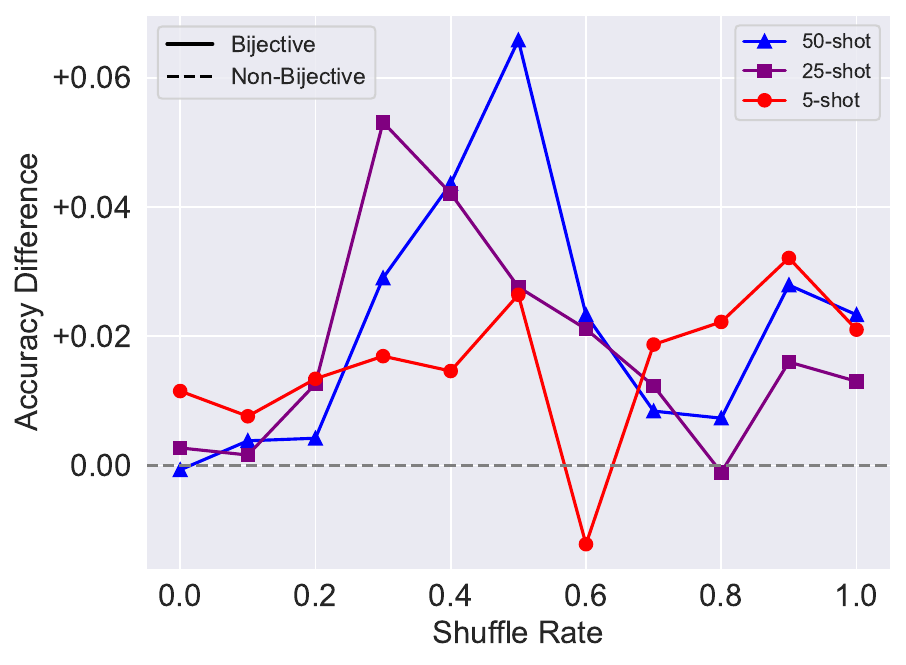}
\caption{
Llama 3.1 8B performance on SST-2 dataset, which shows similar trends with Fig.\ref{fig:amazon-curve}.
\textbf{Left:} 
Under the \textsc{Bijective} cipher, accuracy decreases smoothly as the shuffle rate increases, highlighting the difficulty in interpreting the ciphered text. 
Accuracy also increases with more demonstrations, suggesting the model's ability to solve \textsc{Bijective} cipher. 
\textbf{Right:} $y$-axis shows the accuracy gap between \textsc{Bijective} and \textsc{Non-Bijective} ciphers. 
For very high shuffle rates (e.g, $> 0.7$), the task become very hard to understand and solve (for the model and even humans) as it becomes ill-defined. 
}\label{fig:sst-curve}
\end{figure*}

\begin{figure*}[h]
\centering
\includegraphics[width=0.46\textwidth]{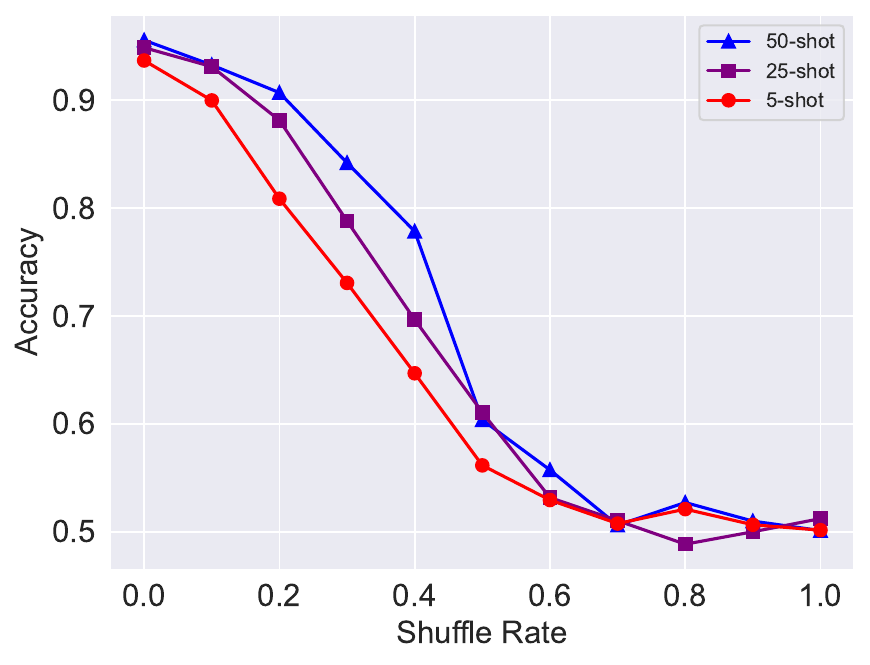} 
\includegraphics[width=0.48\textwidth]{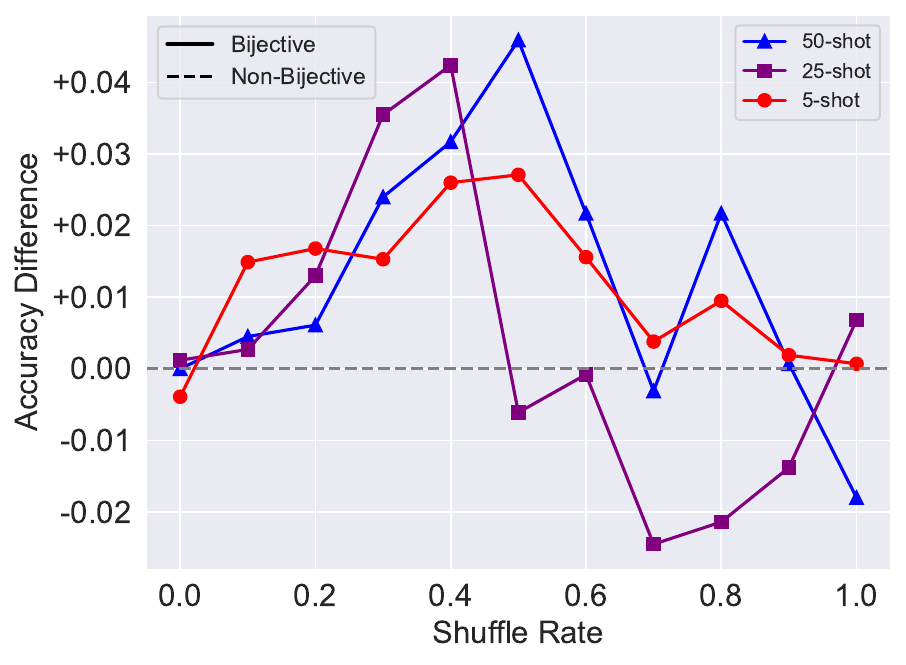}
\caption{
Peformance of Llama 3.1 8B on SST-2 dataset with non-priority sampling, comparing with Fig.\ref{fig:sst-curve}.
\textbf{Left:} The accuracies under \textsc{Bijective} cipher.
\textbf{Right:} The y-axis
displays the accuracy gap between \textsc{Bijective} and \textsc{Non-Bijective} ciphers.
}\label{fig:sst-curve-random}
\end{figure*}

\begin{figure*}[h]
\centering
\includegraphics[width=0.46\textwidth]{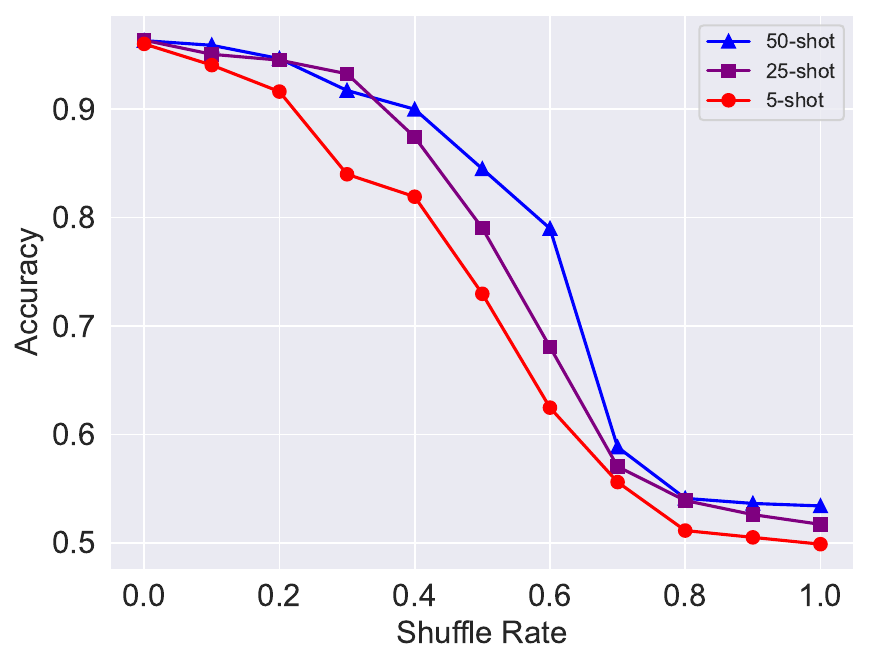}
\includegraphics[width=0.48\textwidth]{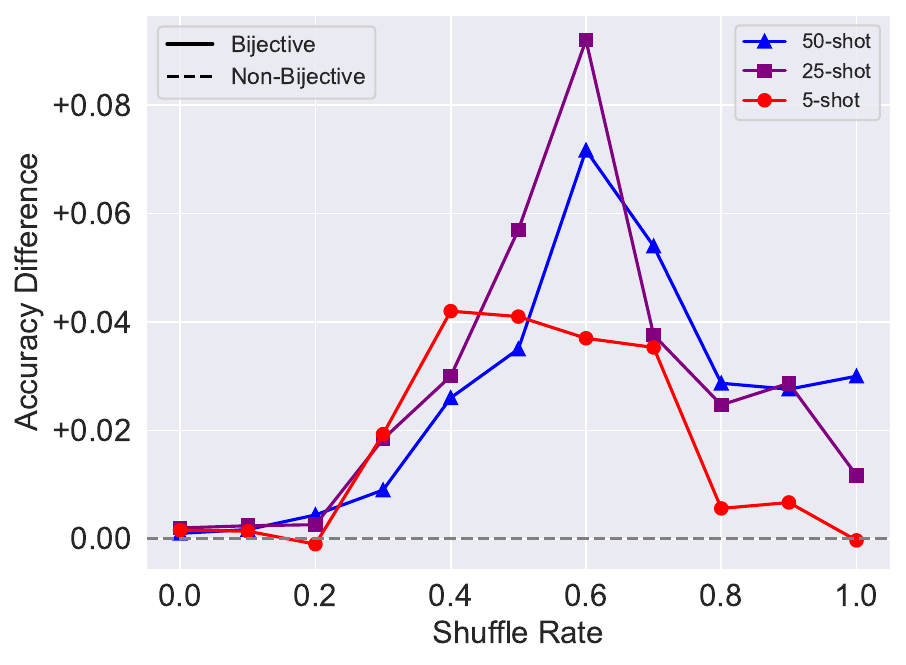}
\caption{
Peformance of Llama 3.1 8B on Amazon dataset with non-priority sampling, comparing with Fig.\ref{fig:amazon-curve}.
\textbf{Left:} The accuracies under \textsc{Bijective} cipher.
\textbf{Right:} The y-axis
displays the accuracy gap between \textsc{Bijective} and \textsc{Non-Bijective} ciphers.
}\label{fig:amazon-curve-random}
\end{figure*}


\clearpage
\section{Pretrained-only vs. Aligned Models}
\label{appendix:ins-model}
\autoref{tab:llama8b_ins_infor_sample} shows the performances of Llama3.1-8B-Instruct on different datasets.
Comparing with its pretrained-only version (\autoref{tab:llama8b_infor_sample}), it demonstrates better performances. 
However, their gaps between \textsc{Bijective} and \textsc{Non-Bijective} ciphers are on par.

{
\setlength{\tabcolsep}{0pt}
\begin{table*}[ht]
    \centering
    \small
\resizebox{\linewidth}{!}{
\begin{tabular}{
P{2.8cm}@{\hskip 0.2cm}
P{2.5cm}@{\hskip 0.2cm}
L{1.8cm}@{\hskip 0.2cm}
L{1.8cm}@{\hskip 0.2cm}
L{1.8cm}@{\hskip 0.2cm}
L{1.8cm}@{\hskip 0.2cm}
L{1.8cm}@{\hskip 0.2cm}
L{1.8cm}@{\hskip 0.2cm}
}
    \toprule
    Shots $\rightarrow$ &\multirow{2}{*}{Cipher}& \multicolumn{6}{c}{Model: Llama 3.1 8B Instruct} \\
     \cmidrule(lr){3-8}
     Dataset (shuffle rate)$\downarrow$ & & 5-shot & 10-shot & 15-shot & 20-shot & 25-shot & 50-shot \\
     \midrule
     \multirow{2}{*}{SST-2 ($r=0.5$)} & \textsc{Non-Bijective} & 65.5  & 66.5   & 68.9   & 68.1  & 67.5  & 65.5     \\
     & \textsc{Bijective} & 69.8   {(\textcolor{DarkGreen}{+4.3} $\uparrow$)}$^*$ & 70.8  {(\textcolor{DarkGreen}{+4.3} $\uparrow$)}$^*$ & 72.4   {(\textcolor{DarkGreen}{+3.5} $\uparrow$)}$^*$ & 70.8   {(\textcolor{DarkGreen}{+2.7} $\uparrow$)}$^*$ & 70.0   {(\textcolor{DarkGreen}{+2.5} $\uparrow$)}$^*$ & 71.8   {(\textcolor{DarkGreen}{+6.3} $\uparrow$)}$^*$ \\
    \midrule
    \multirow{2}{*}{Amazon ($r=0.6$)} & \textsc{Non-Bijective} & 70.5  & 80.0  & 77.3   & 79.3   & 80.6  & 80.5   \\
    & \textsc{Bijective} & 75.8   {(\textcolor{DarkGreen}{+5.3} $\uparrow$)}$^*$ & 82.7   {(\textcolor{DarkGreen}{+2.7} $\uparrow$)}$^*$ & 82.4   {(\textcolor{DarkGreen}{+5.1} $\uparrow$)}$^*$ & 82.4   {(\textcolor{DarkGreen}{+3.1} $\uparrow$)}$^*$ & 84.6   {(\textcolor{DarkGreen}{+4.0} $\uparrow$)}$^*$ & 86.1   {(\textcolor{DarkGreen}{+5.6} $\uparrow$)}$^*$ \\
    \midrule
    
    \multirow{2}{*}{HellaSwag ($r=0.3$)} & \textsc{Non-Bijective} & 43.2   & 43.2   & 42.3   & 41.6  & 41.4  & 41.0     \\
    & \textsc{Bijective} & 44.8      {(\textcolor{DarkGreen}{+1.6} $\uparrow$)}$^*$ & 47.5    {(\textcolor{DarkGreen}{+4.3} $\uparrow$)}$^*$ & 44.4  {(\textcolor{DarkGreen}{+2.1} $\uparrow$)}$^*$ & 44.8     {(\textcolor{DarkGreen}{+3.2} $\uparrow$)} & 45.1    {(\textcolor{DarkGreen}{+3.7} $\uparrow$)}$^*$ & 42.4   {(\textcolor{DarkGreen}{+1.4} $\uparrow$)}$^*$ \\
    \midrule
    
    \multirow{2}{*}{WinoGrande ($r=0.1$)} & \textsc{Non-Bijective} & 57.4  & 58.1   & 55.7  & 57.3   & 56.4  & 57.1   \\
    & \textsc{Bijective} & 59.0    {(\textcolor{DarkGreen}{+1.6} $\uparrow$)} & 58.7   {(\textcolor{DarkGreen}{+0.6} $\uparrow$)}$^*$ & 57.4   {(\textcolor{DarkGreen}{+1.7} $\uparrow$)}$^*$ & 59.3   {(\textcolor{DarkGreen}{+2.0} $\uparrow$)}$^*$ & 58.2  {(\textcolor{DarkGreen}{+1.8} $\uparrow$)}$^*$ & 57.4  {(\textcolor{DarkGreen}{+0.3} $\uparrow$)}$^*$ \\
    \bottomrule
    
    \end{tabular}
    }
\caption{Llama3.1 8B Instruct accuracies (reported in \%) on different datasets with varying numbers of ICL examples under \textsc{Bijective} vs. \textsc{Non-Bijective} ciphers, as comparing to \autoref{tab:llama8b_infor_sample}. 
The numbers inside the parenthesis shows the change from \textsc{Non-Bijective}  to \textsc{Bijective} cipher. 
Statistically significant gains are indicated via $^*$.
}
\label{tab:llama8b_ins_infor_sample}
\end{table*}
}

\clearpage

\section{Small vs. Large Models}
\label{appendix:large-model}

\autoref{tab:llama70b_infor_sample} shows the performances of Llama3.1-70B on different datasets.
Comparing with Llama3.1-8B (\autoref{tab:llama8b_infor_sample}), it demonstrates better performances. 
However, their differences in gaps between \textsc{Bijective} and \textsc{Non-Bijective} ciphers are mixed.

{
\setlength{\tabcolsep}{0pt}
\begin{table*}[ht]
    \centering
    \small
\resizebox{\linewidth}{!}{
\begin{tabular}{
P{2.8cm}@{\hskip 0.2cm}
P{2.5cm}@{\hskip 0.2cm}
L{1.8cm}@{\hskip 0.2cm}
L{1.8cm}@{\hskip 0.2cm}
L{1.8cm}@{\hskip 0.2cm}
L{1.8cm}@{\hskip 0.2cm}
L{1.8cm}@{\hskip 0.2cm}
L{1.8cm}@{\hskip 0.2cm}
}
    \toprule
    Shots $\rightarrow$ &\multirow{2}{*}{Cipher}& \multicolumn{6}{c}{Model: Llama 3.1 70B} \\
     \cmidrule(lr){3-8}
     Dataset (shuffle rate)$\downarrow$ & & 5-shot & 10-shot & 15-shot & 20-shot & 25-shot & 50-shot \\
     \midrule
     \multirow{2}{*}{SST-2 ($r=0.5$)} & \textsc{Non-Bijective} & 64.8   & 70.3   & 66.4   & 71.8    & 69.8   & 74.5     \\
     & \textsc{Bijective} & 68.9   {(\textcolor{DarkGreen}{+4.1} $\uparrow$)}$^*$ & 71.0   {(\textcolor{DarkGreen}{+0.7} $\uparrow$)} & 69.6   {(\textcolor{DarkGreen}{+3.2} $\uparrow$)}$^*$ & 76.9   {(\textcolor{DarkGreen}{+5.1} $\uparrow$)}$^*$ & 74.4   {(\textcolor{DarkGreen}{+4.6} $\uparrow$)}$^*$ & 80.3   {(\textcolor{DarkGreen}{+5.8} $\uparrow$)}$^*$ \\
    \midrule
    \multirow{2}{*}{Amazon ($r=0.6$)} & \textsc{Non-Bijective} & 73.1  & 73.7   & 80.6   & 77.7   & 79.3   & 82.0    \\
    & \textsc{Bijective} & 76.5   {(\textcolor{DarkGreen}{+3.4} $\uparrow$)}$^*$ & 78.6   {(\textcolor{DarkGreen}{+4.9} $\uparrow$)}$^*$ & 84.4   {(\textcolor{DarkGreen}{+3.8} $\uparrow$)}$^*$ & 82.0   {(\textcolor{DarkGreen}{+4.3} $\uparrow$)}$^*$ & 84.0   {(\textcolor{DarkGreen}{+4.7} $\uparrow$)}$^*$ & 85.7   {(\textcolor{DarkGreen}{+3.7} $\uparrow$)}$^*$ \\
    \midrule

    \multirow{2}{*}{HellaSwag ($r=0.3$)} & \textsc{Non-Bijective} & 42.2  & 39.1   & 40.4   & 39.6   & 40.6  & 38.5 \\
    & \textsc{Bijective} & 44.2    {(\textcolor{DarkGreen}{+2.0} $\uparrow$)}$^*$ & 43.6   {(\textcolor{DarkGreen}{+4.5} $\uparrow$)}$^*$ & 43.1   {(\textcolor{DarkGreen}{+2.7} $\uparrow$)}$^*$ & 42.2   {(\textcolor{DarkGreen}{+2.6} $\uparrow$)}$^*$ & 40.9   {(\textcolor{DarkGreen}{+0.3} $\uparrow$)}$^*$ & 41.6 {(\textcolor{DarkGreen}{+3.1} $\uparrow$)}$^*$ \\
    \midrule
    
    \multirow{2}{*}{WinoGrande ($r=0.1$)} & \textsc{Non-Bijective} & 65.1   & 69.5  & 69.9    & 70.1  & 71.0  & 67.4      \\
    & \textsc{Bijective} & 68.6      {(\textcolor{DarkGreen}{+3.5} $\uparrow$)}$^*$ & 70.1    {(\textcolor{DarkGreen}{+0.6} $\uparrow$)} & 71.2   {(\textcolor{DarkGreen}{+1.3} $\uparrow$)}$^*$ & 71.4     {(\textcolor{DarkGreen}{+1.3} $\uparrow$)}$^*$ & 72.2    {(\textcolor{DarkGreen}{+1.2} $\uparrow$)}$^*$ & 70.8    {(\textcolor{DarkGreen}{+3.4} $\uparrow$)}$^*$ \\
    
    \bottomrule
    \end{tabular}
    }
    \caption{Llama3.1 70B accuracies (reported in \%) on different datasets with varying numbers of ICL examples under \textsc{Bijective} vs. \textsc{Non-Bijective} ciphers. 
    The numbers inside the parenthesis shows the change from \textsc{Non-Bijective}  to \textsc{Bijective} cipher. 
    Statistically significant gains are indicated via $^*$.
    }
    \label{tab:llama70b_infor_sample}
\end{table*}
}

\section{Statistical Significance of Results}
\label{appendix:sig}
To determine if the gaps between \textsc{Bijective} and \textsc{Non-Bijective} ciphers are significant, we conduct McNemar's test \cite{McNemar_1947}. 
\autoref{tab:sig_fixed_shot_r}, \autoref{tab:sig_llama8b_infor_sample}, \autoref{tab:sig_llama8b_ins_infor_sample} and \autoref{tab:sig_llama70b_infor_sample} show the computed p-values for \autoref{tab:fixed_shot_r}, \autoref{tab:llama8b_infor_sample}, \autoref{tab:llama8b_ins_infor_sample} and \autoref{tab:llama70b_infor_sample} 
respectively.
The gap is regard as significant if its corresponding p-value is no larger than $0.05$.

\begin{table*}[h]
    \centering
    \small
\begin{tabular}{
P{3cm}@{\hskip 0.2cm}
P{1.9cm}@{\hskip 0.2cm}
P{1.9cm}@{\hskip 0.2cm}
P{1.9cm}@{\hskip 0.2cm}
P{1.9cm}@{\hskip 0.2cm}
}
    \toprule
    Model $\rightarrow$ & \multicolumn{4}{c}{ 20-shot  } \\
     \cmidrule(lr){2-5}
     Dataset (shuffle rate) $\downarrow$ & Llama3.1 & Qwen2.5 & Olmo & Gemma2\\
     \midrule
    SST-2 ($r=0.5$)  & \textbf{0.000}   & \textbf{0.000}  & \textbf{0.000}  & \textbf{0.001}   \\
    \midrule
    
    Amazon ($r=0.6$) & \textbf{0.000}   & \textbf{0.000}   & \textbf{0.000}   & \textbf{0.000}     \\ 
    \midrule
    
    HellaSwag ($r=0.3$) & \textbf{0.000}  & \textbf{0.000}  & \textbf{0.000}   &  0.663     \\
    \midrule
    
    WinoGrande ($r=0.1$)  & \textbf{0.000}   & 0.084   & 0.786   & 0.943     \\
    \bottomrule
    
    \end{tabular}
    \caption{Significance results (p-values) of McNemar's test for \autoref{tab:fixed_shot_r}. 
    The gap between \textsc{Bijective} and \textsc{Non-Bijective} can be regared as significant if its corresponding p-value is no larger than $0.05$, which is bolded.
    }
    \label{tab:sig_fixed_shot_r}
\end{table*}
{
\setlength{\tabcolsep}{0pt}
\begin{table*}[h]
    \centering
    \small
\resizebox{\linewidth}{!}{
\begin{tabular}{
P{3cm}@{\hskip 0.2cm}
P{1.7cm}@{\hskip 0.2cm}
P{1.7cm}@{\hskip 0.2cm}
P{1.7cm}@{\hskip 0.2cm}
P{1.7cm}@{\hskip 0.2cm}
P{1.7cm}@{\hskip 0.2cm}
P{1.7cm}@{\hskip 0.2cm}
}
    \toprule
    Shots $\rightarrow$ & \multicolumn{6}{c}{Model: Llama 3.1 8B} \\
     \cmidrule(lr){2-7}
     Dataset (shuffle rate)$\downarrow$ & 5-shot & 10-shot & 15-shot & 20-shot & 25-shot & 50-shot \\
     \midrule
     SST-2 ($r=0.5$) & \textbf{0.018}   & 0.205  & 0.084   & \textbf{0.000}  & \textbf{0.020}   & \textbf{0.000}     \\
    \midrule
    Amazon ($r=0.6$) & \textbf{0.000}   & \textbf{0.000}  & \textbf{0.000}   & \textbf{0.000}   & \textbf{0.000}   & \textbf{0.000}   \\
    \midrule
    
    HellaSwag ($r=0.3$)  & \textbf{0.015}   & 0.627   & \textbf{0.000}   & \textbf{0.000}    & 0.278    & 0.357      \\
    \midrule
    
    WinoGrande ($r=0.1$)  & 0.110    & \textbf{0.000}   & \textbf{0.000}   & \textbf{0.000}    & \textbf{0.000}   & \textbf{0.000}    \\
    \bottomrule
    \end{tabular}
    }
    \caption{Significance results (p-values) of McNemar's test for \autoref{tab:llama8b_infor_sample}. 
    The gap between \textsc{Bijective} and \textsc{Non-Bijective} can be regared as significant if its corresponding p-value is no larger than $0.05$, which is bolded. 
    }
    \label{tab:sig_llama8b_infor_sample}
\end{table*}
}
{
\setlength{\tabcolsep}{0pt}
\begin{table*}[h]
    \centering
    \small
\resizebox{\linewidth}{!}{
\begin{tabular}{
P{3cm}@{\hskip 0.2cm}
P{1.7cm}@{\hskip 0.2cm}
P{1.7cm}@{\hskip 0.2cm}
P{1.7cm}@{\hskip 0.2cm}
P{1.7cm}@{\hskip 0.2cm}
P{1.7cm}@{\hskip 0.2cm}
P{1.7cm}@{\hskip 0.2cm}
}
    \toprule
    Shots $\rightarrow$ & \multicolumn{6}{c}{Model: Llama 3.1 8B Instruct} \\
     \cmidrule(lr){2-7}
     Dataset (shuffle rate)$\downarrow$ & 5-shot & 10-shot & 15-shot & 20-shot & 25-shot & 50-shot \\
     \midrule
     SST-2 ($r=0.5$) & \textbf{0.000}  & \textbf{0.000}  & \textbf{0.001}   & \textbf{0.016}  & \textbf{0.025}   & \textbf{0.000}     \\
    \midrule
    Amazon ($r=0.6$) & \textbf{0.003}   & \textbf{0.002}  & \textbf{0.000}   & \textbf{0.000}   & \textbf{0.000}   & \textbf{0.000}    \\
    \midrule
    
    HellaSwag ($r=0.3$)   & \textbf{0.000}   & \textbf{0.000}  & \textbf{0.031}   & 0.081    & \textbf{0.000}   & \textbf{0.023}      \\
    \midrule
    
    WinoGrande ($r=0.1$)  & 0.067    & \textbf{0.000}   & \textbf{0.000}   & \textbf{0.013}     & \textbf{0.015}   & \textbf{0.000}    \\
    \bottomrule
    \end{tabular}
    }
    \caption{Significance results (p-values) of McNemar's test for \autoref{tab:llama8b_ins_infor_sample}. 
    The gap between \textsc{Bijective} and \textsc{Non-Bijective} can be regared as significant if its corresponding p-value is no larger than $0.05$, which is bolded. 
    }
    \label{tab:sig_llama8b_ins_infor_sample}
\end{table*}
}
{
\setlength{\tabcolsep}{0pt}
\begin{table*}[h]
    \centering
    \small
\resizebox{\linewidth}{!}{
\begin{tabular}{
P{3cm}@{\hskip 0.2cm}
P{1.7cm}@{\hskip 0.2cm}
P{1.7cm}@{\hskip 0.2cm}
P{1.7cm}@{\hskip 0.2cm}
P{1.7cm}@{\hskip 0.2cm}
P{1.7cm}@{\hskip 0.2cm}
P{1.7cm}@{\hskip 0.2cm}
}
    \toprule
    Shots $\rightarrow$ & \multicolumn{6}{c}{Model: Llama 3.1 70B} \\
     \cmidrule(lr){2-7}
     Dataset (shuffle rate)$\downarrow$ & 5-shot & 10-shot & 15-shot & 20-shot & 25-shot & 50-shot \\
     \midrule
     SST-2 ($r=0.5$) & \textbf{0.000}  & 0.497 & \textbf{0.006}   & \textbf{0.000}  & \textbf{0.000}    & \textbf{0.000}    \\
    \midrule
    Amazon ($r=0.6$) & \textbf{0.000}  & \textbf{0.000} & \textbf{0.000}  & \textbf{0.000}  & \textbf{0.000}  & \textbf{0.000}    \\
    \midrule
    
    HellaSwag ($r=0.3$)  & \textbf{0.000}  & \textbf{0.006}   & \textbf{0.000}  & \textbf{0.000}  & \textbf{0.000}   & \textbf{0.000}    \\
    \midrule
    
    WinoGrande ($r=0.1$)  & \textbf{0.000}  & 0.446  & \textbf{0.000}  & \textbf{0.000}  & \textbf{0.000}  & \textbf{0.000}   \\
    \bottomrule
    \end{tabular}
    }
    \caption{Significance results (p-values) of McNemar's test for \autoref{tab:llama70b_infor_sample}. 
    The gap between \textsc{Bijective} and \textsc{Non-Bijective} can be regared as significant if its corresponding p-value is no larger than $0.05$, which is bolded. 
    }
    \label{tab:sig_llama70b_infor_sample}
\end{table*}
}

\clearpage
\section{Further Results on Probing Analysis}

\
To get a clearer vision, we extract the rank difference from the last layer on SST-2, dividing them equally into 5 chunks, as shown in Fig.\ref{fig:chunks}. 
For random substitution, there is not much change for rank difference. 
For \textsc{Bijective} substitution, rank difference increases as the chunk number gets bigger. 
This suggests that as LLM sees more occurrences of the substitution token, it learns to use the substitution token as the original token, namely solving \name.

\begin{figure*}[h]
\centering
\includegraphics[width=0.45\textwidth]{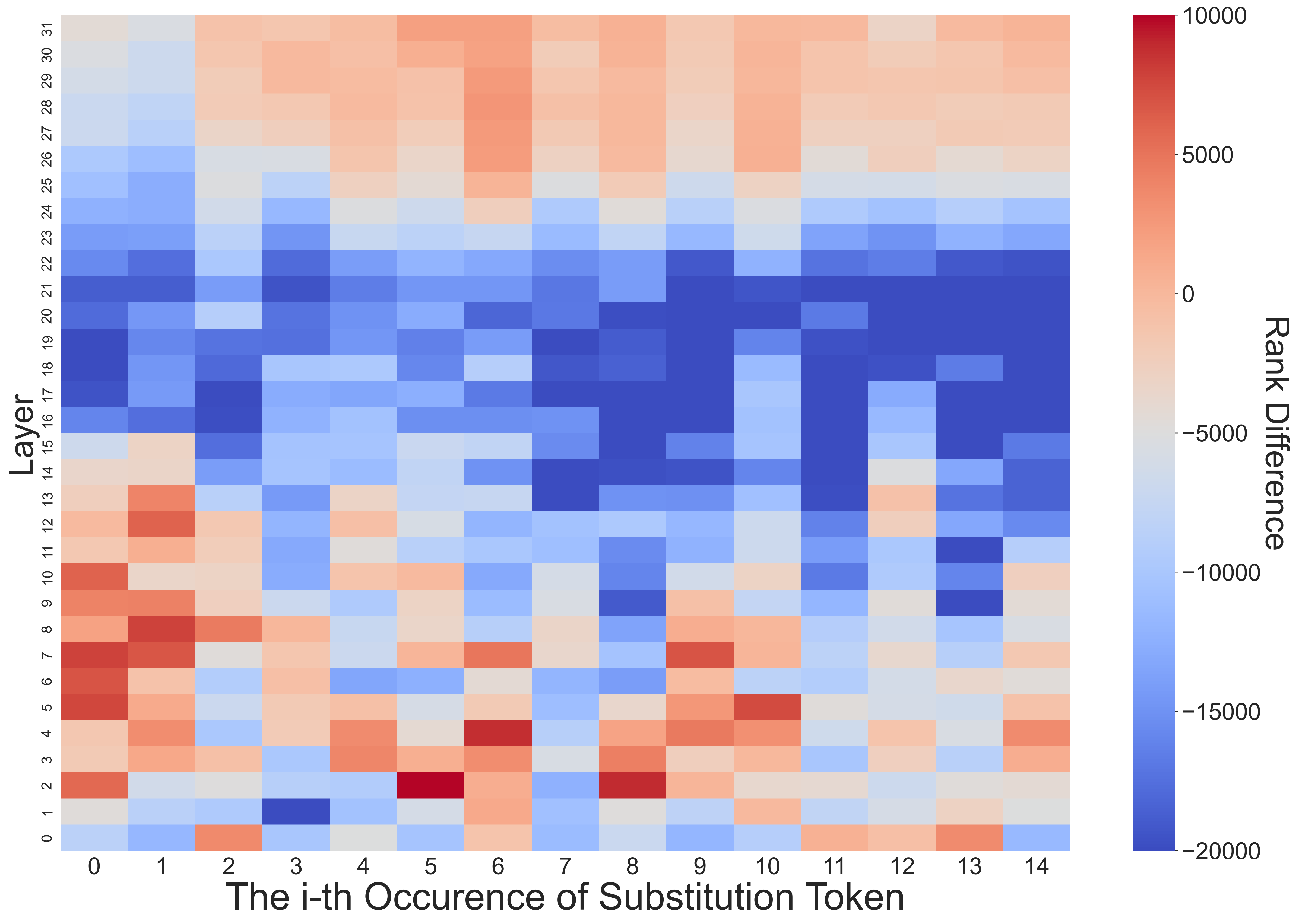} 
\includegraphics[width=0.45\textwidth]{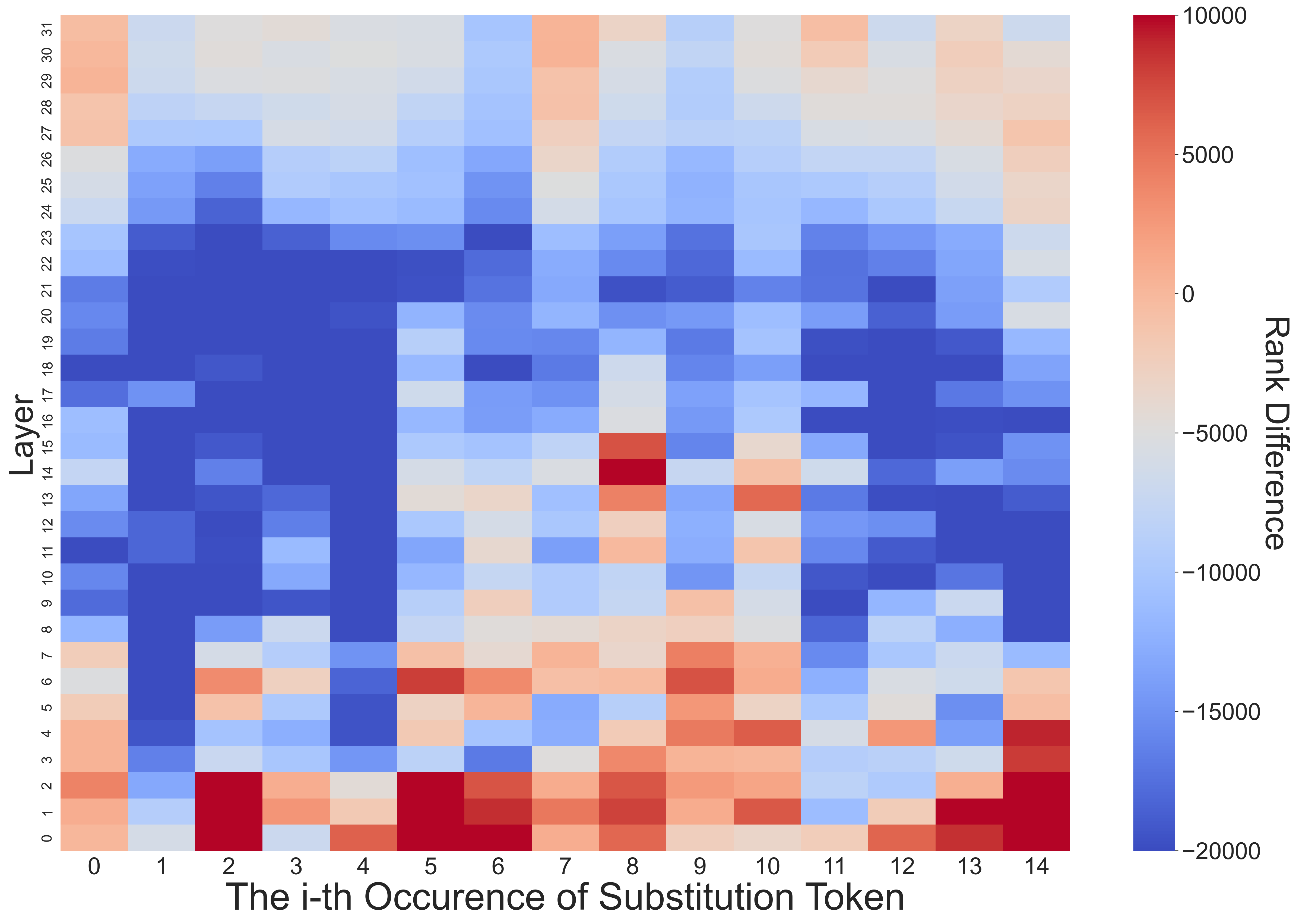}
\caption{Complete heatmap of original token rank minus substitution token rank on Amazon for Fig.\ref{fig:heatmap_half}. \textbf{Left:} \textsc{bijective} cipher \textbf{Right:} \textsc{non-bijective} cipher
}\label{fig:heatmap_whole}
\end{figure*}

\begin{figure*}[h]
\centering
\includegraphics[width=0.48\textwidth]{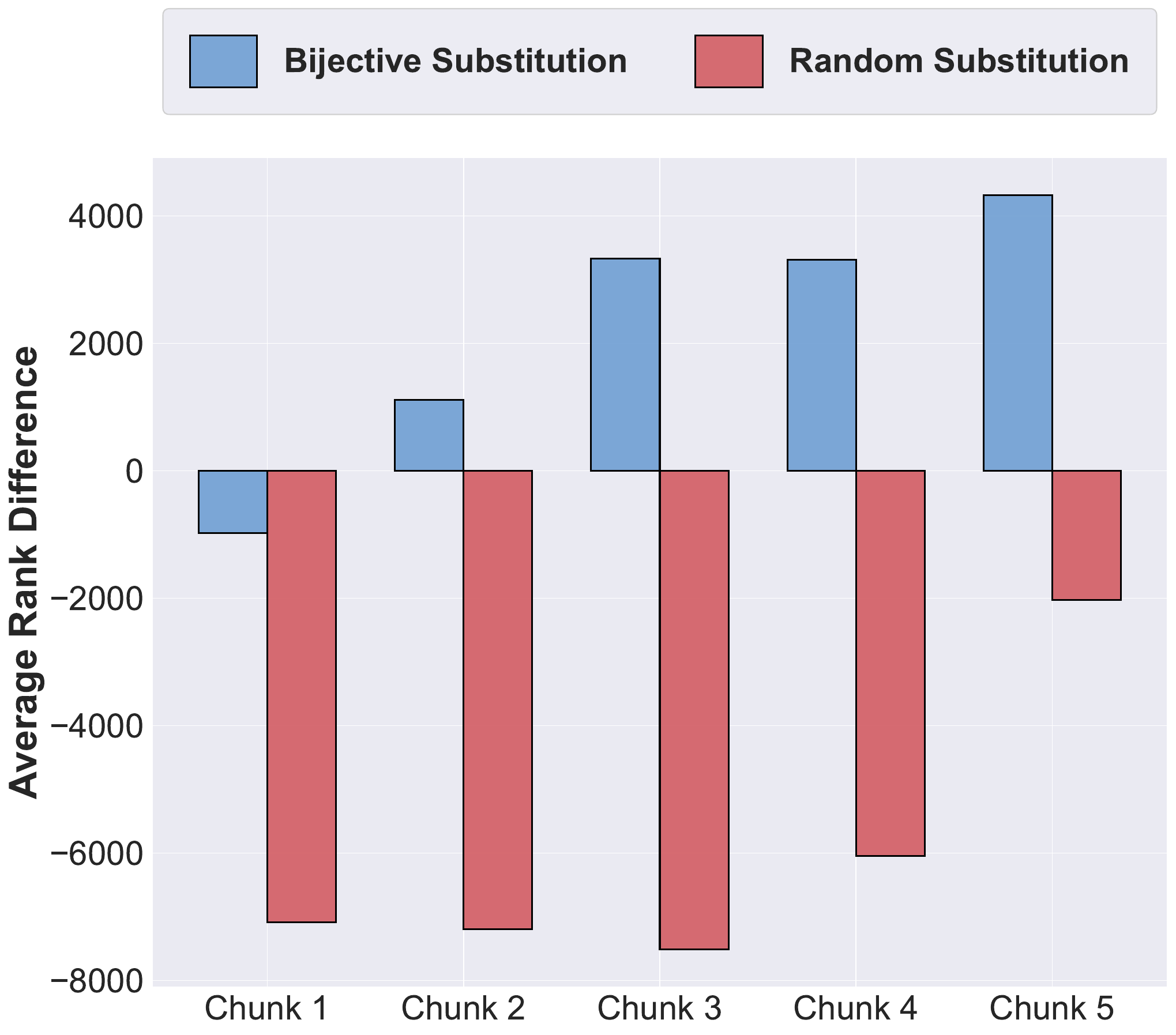}
\includegraphics[width=0.48\textwidth]{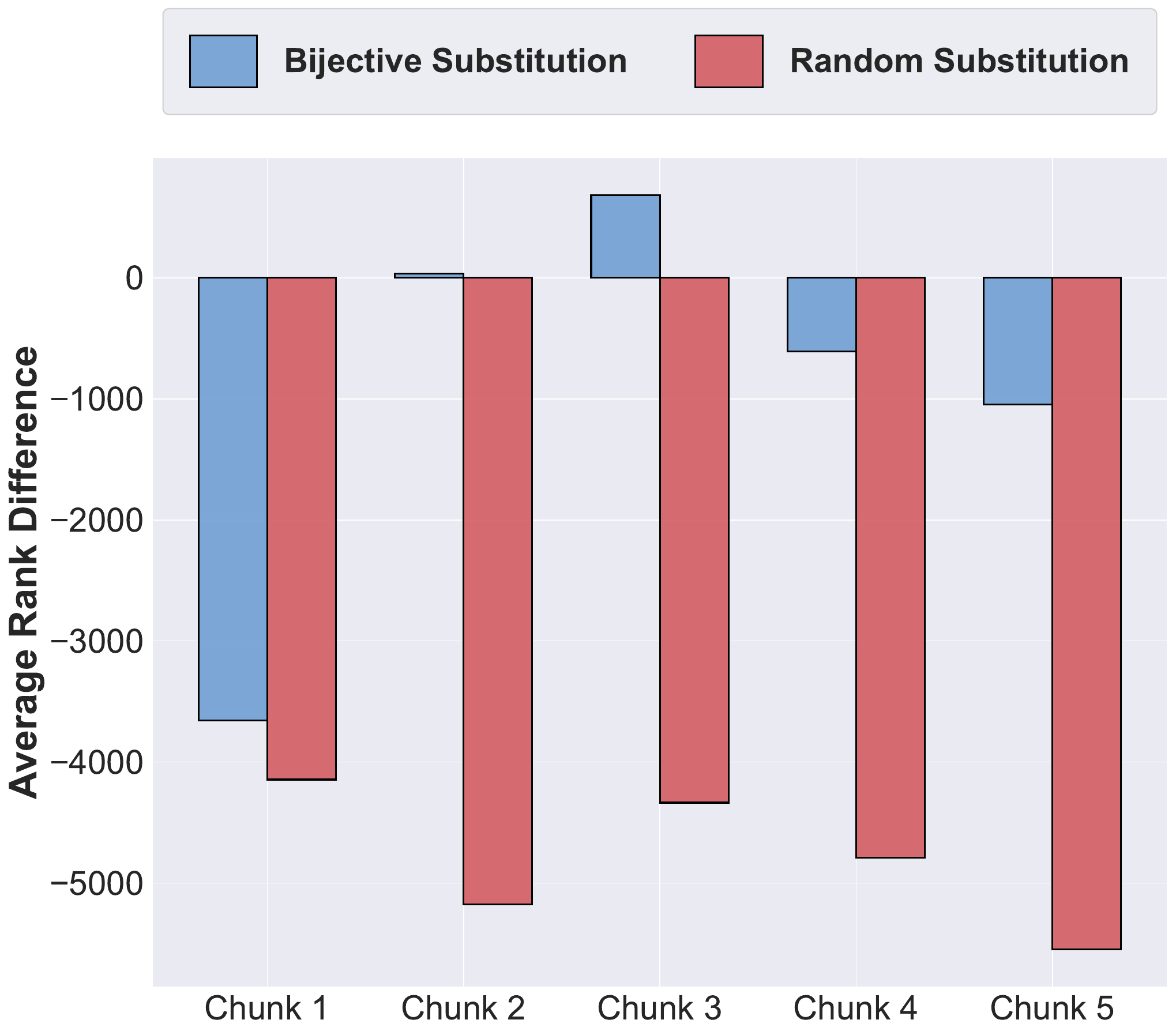}
\caption{
Average rank differences (original token rank - substitution token rank) in SST-2 (left) and Amazon (right) datasets for \textsc{Bijective} (blue) and \textsc{Non-bijective} (red) cipher over 15 occurrences, divided into 5 chunks of size 3. Rank difference serves as a proxy for the model's deciphering ability. Under \textsc{Bijective} cipher, this ability improves with more exposure to substituted tokens, while \textsc{Non-Bijective} cipher shows no clear pattern.
}\label{fig:chunks}
\end{figure*}

\end{document}